\documentclass[runningheads]{llncs}

\usepackage[T1]{fontenc}
\usepackage{graphicx}
\usepackage{subcaption}
\usepackage{comment}
\usepackage{amsmath}
\usepackage{multirow}
\usepackage{arydshln}
\usepackage{amssymb}
\usepackage{changepage}

\begin{document}

\title{MUFASA: Multi-View Fusion and Adaptation Network with Spatial Awareness for Radar Object Detection}
\titlerunning{MUFASA}
%\author{Anonymous submission}

\author{Xiangyuan Peng\inst{1,2} \and
Miao Tang\inst{3} \and
Huawei Sun\inst{1,2}\and
Kay Bierzynski\inst{2}\and
Lorenzo Servadei\inst{1} \and
Robert Wille\inst{1}
}
\authorrunning{X. Peng et al.}
\institute{Technical University of Munich, Chair for Design Automation, Munich, Germany \and
Infineon Technical University of Munich, Neubiberg, Germany
\email{Xiangyuan.Peng@infineon.com} \and
China University of Geosciences (Wuhan), China\\
}

\maketitle   
\begin{abstract}
In recent years, approaches based on radar object detection have made significant progress in autonomous driving systems due to their robustness under adverse weather compared to LiDAR. However, the sparsity of radar point clouds poses challenges in achieving precise object detection, highlighting the importance of effective and comprehensive feature extraction technologies.
To address this challenge, this paper introduces a comprehensive feature extraction method for radar point clouds. 
This study first enhances the capability of detection networks by using a plug-and-play module, GeoSPA. It leverages the Lalonde features to explore local geometric patterns. Additionally, a distributed multi-view attention mechanism, DEMVA, is designed to integrate the shared information across the entire dataset with the global information of each individual frame.
By employing the two modules, we present our method, MUFASA, which enhances object detection performance through improved feature extraction. The approach is evaluated on the VoD and TJ4DRaDSet datasets to demonstrate its effectiveness. In particular, we achieve state-of-the-art results among radar-based methods on the VoD dataset with the mAP of 50.24\%.

\end{abstract}

\section{Introduction}
Object detection plays an important role in autonomous driving systems, providing essential perception capabilities. Initially, object detection relied heavily on camera-based methods, leveraging high-resolution RGB images to accomplish 2D detection \cite{kaur2022tools}. However, cameras lack 3D geometric clues, which is crucial for accurate perception. Hence, Light Detection and Ranging (LiDAR) sensors are integrated into systems to generate 3D point clouds, which are capable of offering geometric details and lead to more accurate object detection \cite{wu2020deep}. Nevertheless, LiDAR's performance diminishes under adverse weather conditions, including fog, rain, and snow \cite{tang2020performance}. The high cost also limits its widespread application \cite{raj2020survey}.

As an alternative, radar has gained sufficient interest due to its robust performance under adverse conditions and capability to provide 3D spatial and velocity information \cite{jiang2022t}. Yet, radar point clouds pose a unique challenge: sparsity. Numerous methods have been proposed to extract rich features from the sparse points, including sparse theory and advanced signal processing techniques \cite{ding2023sparsity,sun20214d}.
However, these methods are still insufficient to fully explore the information in radar point clouds. 

Currently, most feature extraction technologies for sparse radar point clouds primarily focus on extracting local and global features within individual frames \cite{meyer2021graph,yang2020radarnet}. 
While understanding the global semantic information within each frame is essential for scene comprehension, the hidden shared information across the entire dataset is often neglected. For instance, the locations of cars and pedestrians are usually in the center and at the edges of different scenes, respectively. These dataset-wide features can enhance detection confidence by incorporating shared information across different frames.
Moreover, local feature extraction often overlooks the significance of geometric spatial patterns. Accurately distinguishing between similar objects, such as pedestrians and signposts, or cars and square trash bins, heavily relies on precise analysis of local geometric patterns. However, during the learning process in feature extraction, the details of these patterns can be missed.
These considerations underscore the need for more comprehensive and sophisticated feature extraction strategies.

Therefore, an advanced feature extraction network, MUFASA, is proposed to leverage complex local and global information within each radar point cloud frame, along with shared dataset-wide features. 

\textbf{Contribution} This study focuses on enhancing radar point cloud object detection capabilities through comprehensive feature extractions. The main contributions are as follows:
\vspace{-2mm}
\begin{itemize}
\item[$\bullet$] A flexible plug-and-play Geometric Spatial Pattern Analysis module, GeoSPA, is designed to explore local geometric features within various detection networks. It enhances the extraction of local features and provides a deep understanding of spatial information in the feature extraction backbone.
\item[$\bullet$] A Distributed External Multi-View Attention module, DEMVA, is designed to integrate dataset-wide features across different frames. It merges the dataset-wide information with the global features of each frame from the cylinder and Bird's Eye View (BEV) projections.
\item[$\bullet$] Extensive experiments on the View-of-Delft (VoD) \cite{palffy2022multi} and TJ4DRadSet \cite{zheng2022tj4dradset} datasets
demonstrate the superior efficiency and accurate detection performance of the proposed method.
\end{itemize}

\section{Related Work}

\subsection{Radar-based 3D Object Detection}
Radar point cloud detection methods, often inspired by LiDAR techniques, are classified into voxel-based, pillar-based, point-based, and fusion-based networks \cite{mao20223d}.
Voxel-based strategies \cite{shi2019part,sindagi2019mvx} convert point clouds into 3D voxel grids, sacrificing some information due to discretization.
Pillar-based approaches, such as FastPillars \cite{zhou2023fastpillars}, transform point clouds into 2D pseudo-images for BEV analysis, effectively utilizing 2D convolutional neural networks (CNNs) for feature extraction.
Point-based methods, highlighted by PointNet \cite{qi2017pointnet}, directly extract features from raw points, preserving fine-grained details. However, they may suffer higher computational costs.
Fusion-based strategies, like PV-RCNN \cite{shi2020pv}, integrate multiple approaches to improve detection performance while considering computational demands.

Despite advancements, extracting precise features from radar point clouds with a single method remains a significant challenge because of the sparsity. Therefore, our network integrates pillar and point-based approaches to ensure comprehensive feature extraction while balancing computational efficiency and detection accuracy.

\subsection{Descriptors for Feature Extraction}
The inherent sparsity and irregular distribution of radar point clouds necessitate efficient feature extraction in 3D spaces. An effective point cloud descriptor is capable of capturing spatial geometric characteristics while being invariant to translation, scaling, and rotation. Optimizing these descriptors can improve object detection accuracy while reducing the computational cost.

Conventional descriptors combine spatial statistical information of feature points and their neighbors to extract geometric features \cite{lalonde2006natural,ghorbani2022novel}. With the advancement of deep learning, integrating hand-crafted and deep learning-derived features has marked a significant evolution. For instance, \cite{khoury2017learning} employs spherical histograms for spatial mapping around points, with neural networks transforming these representations into Euclidean space. PPFNet \cite{deng2018ppfnet} and PPF-FoldNet \cite{deng2018ppf} leverage Fisher Vectors in CNNs to navigate the unstructured characteristic of point clouds. 
The fusion of traditional 3D feature descriptors with deep learning approaches represents a noteworthy advancement in point cloud processing, resulting in more robust and invariant feature representations.

However, a major drawback of most feature extraction methods is the descriptors are tightly coupled with specific network structures, limiting their applicability across different detection systems. To address this limitation, we introduce our GeoSPA, which can integrate into various detection networks, enhancing the generalization of feature extraction.

\subsection{Attention Mechanism in 3D Object Detection}
The attention mechanism achieves permutation invariance, dynamically focuses on relevant features, and adapts to sparsity and density variations. Also, the ability of attention mechanisms to stably extract information makes them particularly suitable for the complex point cloud analysis \cite{qiu2021investigating}.

For instance, Box attention \cite{nguyen2022boxer} introduces a grid-based attention weighting for 3D object detection in outdoor scenes. TANet \cite{liu2020tanet} enhances point cloud discrimination by applying attention at multiple levels to refine the feature representation. CSANet \cite{wang2022cross} employs self-attention to fuse point cloud features with coordinate information. Additionally, Point-BERT \cite{yu2022point} designs a transformer-based pre-training for point clouds.

These advancements, mainly based on the self-attention mechanism, highlight the flexibility and effectiveness of the attention mechanism in 3D object detection. However, external attention, which has a strong learning ability to integrate the information across the entire dataset, is not commonly employed in point cloud detection. Therefore, we designed DEMVA to capture the shared dataset-wide information.
\section{Proposed Method}
In this section, we outline the overall architecture of MUFASA. Subsequently, the GeoSPA module, designed to extract local geometric patterns within the point clouds, is presented. Later, the DEMVA module is introduced, which utilizes shared information across the entire dataset and the global features in each frame, thereby enhancing the model's overall performance and robustness.

\subsection{Overall Structure}
\begin{figure}[h]
    \centering
    \includegraphics[width=1\linewidth]{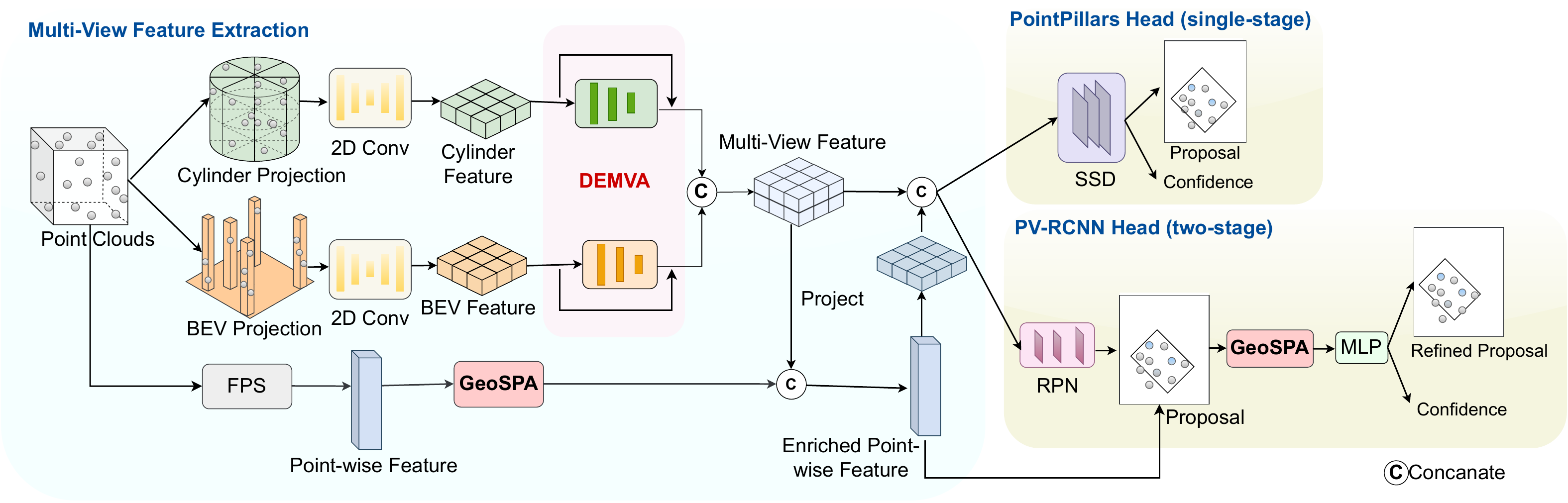}
    \caption{Overall Structure of MUFASA} 
    \label{fig: network structure}
    \vspace{-6mm}
\end{figure}

The structure of MUFASA is illustrated in Fig. \ref{fig: network structure}.
Initially, the radar point clouds undergo transformation into multi-view representations, including BEV and cylindrical projections, inspired by the MVFAN \cite{yan2023mvfan}. On the one hand, extracted features through the cylindrical branch maintain the integrity of 3D information. On the other hand, the BEV branch captures a comprehensive scene perspective, effectively mitigating obstruction by obstacles.
Subsequently, the features from both representations are fed into the DEMVA. The distributed attention mechanisms enhance feature comprehension by exploring relationships across the entire dataset, addressing and mitigating potential information loss.

In parallel, our network incorporates a point-wised branch with Farthest Point Sampling (FPS) and our GeoSPA module, designed to capture point-wise features. The point-wise features are then fused with the multi-view features derived from the DEMVA branch.
Enriched point-wise features are subsequently reprojected back into the BEV. The final features are treated as input for the detection heads to generate proposals. 

The GeoSPA is designed for integration with various detection frameworks, acting as a flexible plug-and-play module integrated into point-wise feature extraction. In single-stage detectors like PointPillars \cite{lang2019pointpillars}, with a detection head SSD \cite{liu2016ssd}, GeoSPA enriches point-wise feature extraction without the need for proposal refinement. In two-stage detectors like PV-RCNN \cite{shi2020pv}, where the proposal generation head is RFN \cite{shi2020pv},  GeoSPA can be utilized in two aspects. It is applied not only in the point-wise feature extraction but also within the Region of Interest (RoI) to refine spatial features, aiding in the proposal refinement. The implementation of GeoSPA in different locations demonstrates its adaptability across different detection architectures as a plug-and-play module.

\subsection{Geometric Spatial Pattern Analysis}

In road object detection, cars often exhibit distinct rectangular shapes, particularly for stationary vehicles, where Doppler velocity provides less information. Meanwhile, cyclists differ from pedestrians in both width and height. To enhance the features obtained from deep learning alone, MUFASA integrates the GeoSPA based on the Lalonde descriptor, which is able to capture spatial features.
\vspace{-1mm}
\paragraph{Definition of Lalonde Feature:} The spatial descriptors are derived through two processes: Signature and Histogram. Following the methodology outlined in \cite{lalonde2006natural}, we analyze the distribution of points within a local spatial domain by conducting Principal Component Analysis (PCA) on the covariance matrix of the 3D points' locations. The covariance matrix \(M\) is expressed as Equation \ref{eq: coveriance}
\vspace{-3mm}
\begin{equation}
M = \frac{1}{N} \sum_{i=1}^{N} (X_i - \bar{X})(X_i - \bar{X})^T,
\label{eq: coveriance}
\vspace{-3mm}
\end{equation}
where $\{X_i\} = \{(x_i, y_i, z_i)^T\}$ is the $i^{th}$ point in the point clouds and $\bar{X} = \frac{1}{N} \sum_{i=1}^{N} X_i$ is the mean value of all the points. The eigenvectors obtained from PCA are $e_1$, $e_2$ and $e_3$ with eigenvalus $x_1$, $x_2$ and $x_3$ while $x_1 \geq x_2 \geq x_3$. For different geometric representations, the eigenvalues showcase different features:
\vspace{-2mm}
\begin{align*} 
x_1 \approx x_2 \approx x_3 &  \text{  for scattered points}, \\
x_1 \gg x_2 \approx x_3 &  \text{  for a linear structure}, \\
x_1 \approx x_2 \gg x_3 &  \text{  for a solid surface}.
\vspace{-10mm}
\end{align*}
Then, $l_{scatter}$, $l_{linear}$, and $l_{surface}$ are defined as the scatter-ness, linear-ness, and surface-ness. They can be calculated as Equation \ref{eq:my_cases}
\vspace{-2mm}
\begin{equation}
l_{scatter} = x_1, \quad l_{linear} = x_1 - x_2, \quad l_{surface} = x_2 - x_3.
\label{eq:my_cases}
\vspace{-2mm}
\end{equation}
By taking the histogram of $l_{scatter}$, $l_{linear}$, and $l_{surface}$ for all points within the point clouds, we obtain the three Lalonde features $L_1$, $L_2$, and $L_3$ of the object.
\begin{figure}
    \vspace{-4mm}
    \centering
    \begin{subfigure}[b]{0.3\textwidth}
        \includegraphics[width=\textwidth]{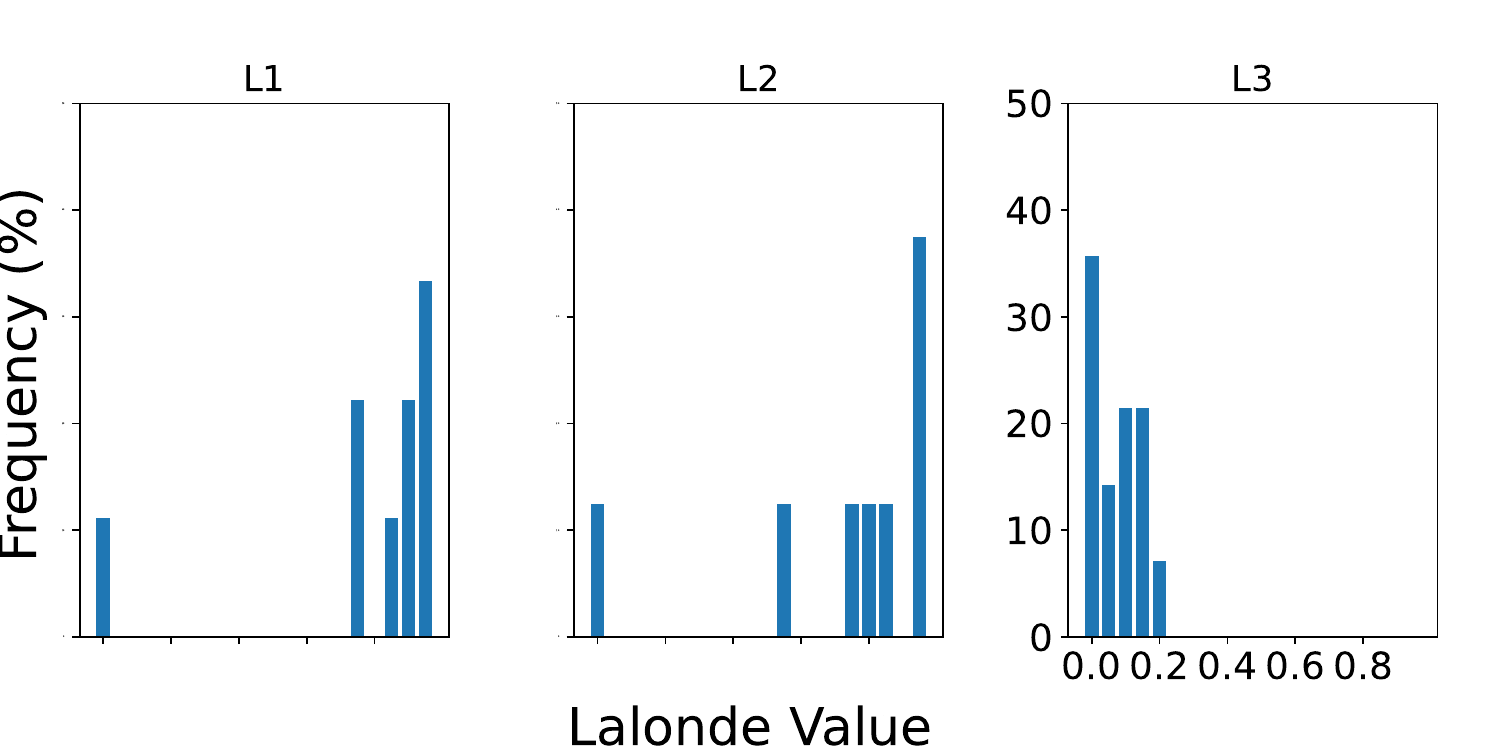}
        \caption{Car side to radar.}
        \label{fig: Car0-10}
    \end{subfigure}
        \begin{subfigure}[b]{0.3\textwidth}
        \includegraphics[width=\textwidth]{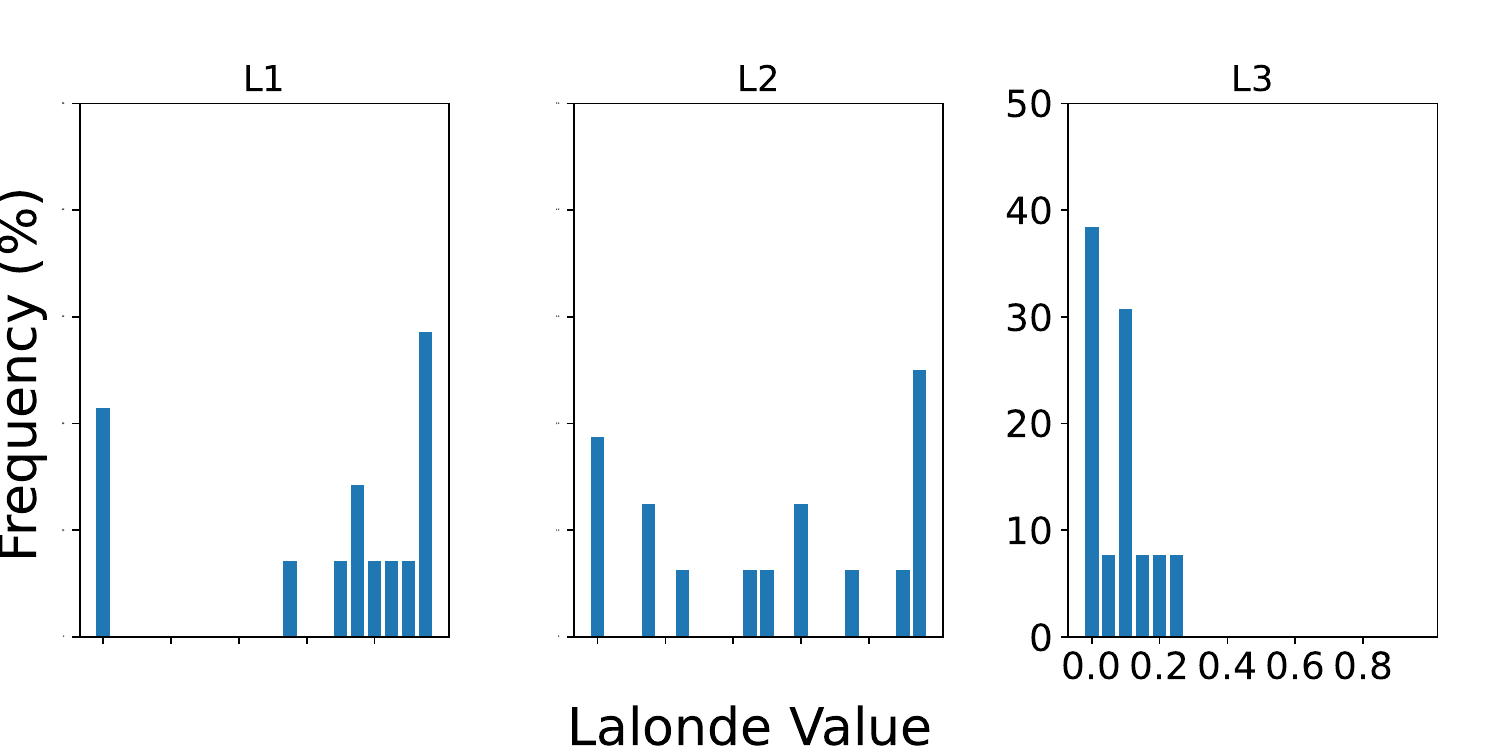}
        \caption{Car skew to radar.}
        \label{fig: Car40-50}
    \end{subfigure}
        \begin{subfigure}[b]{0.3\textwidth}
        \includegraphics[width=\textwidth]{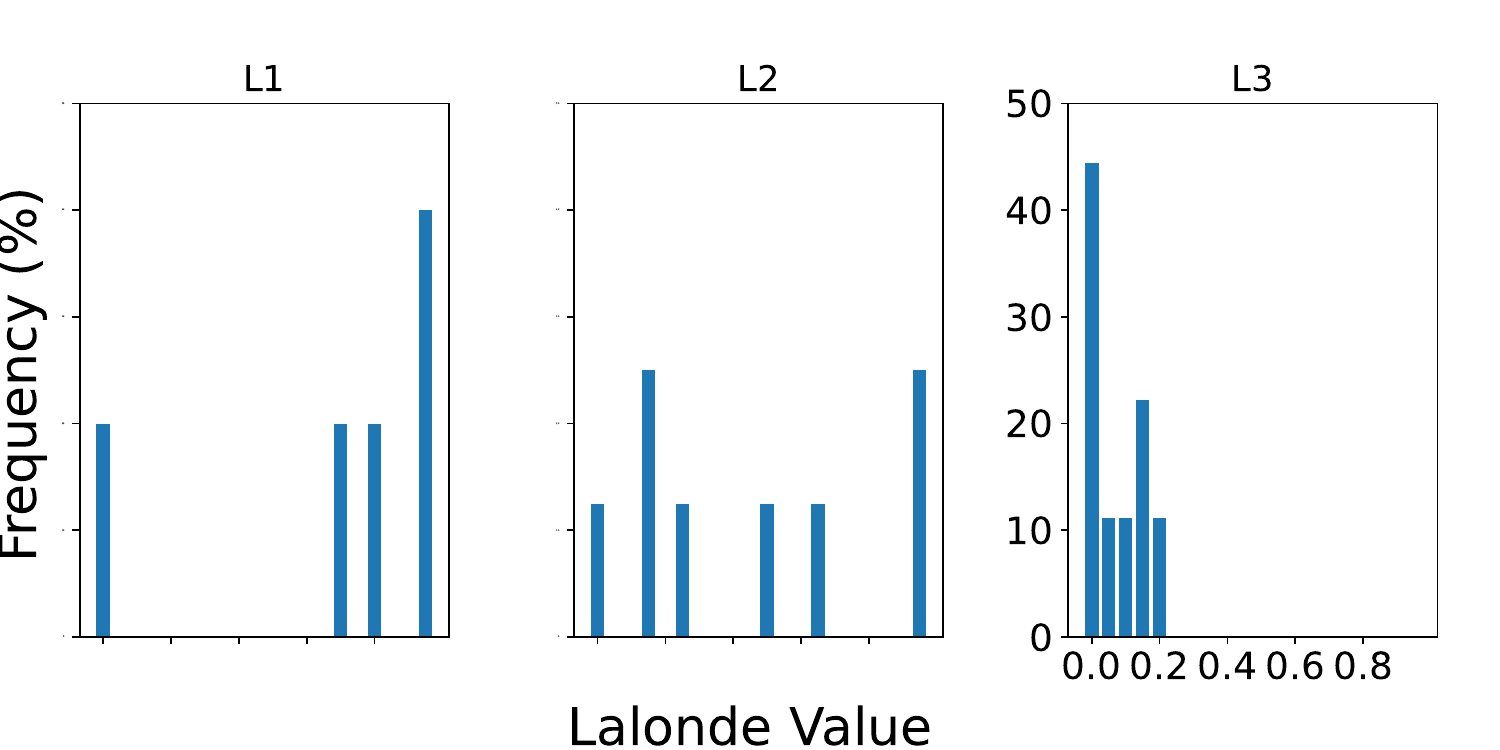}
        \caption{Car face to radar.}
        \label{fig: Car80-90}
    \end{subfigure}
    \begin{subfigure}[b]{0.3\textwidth}
        \includegraphics[width=\textwidth]{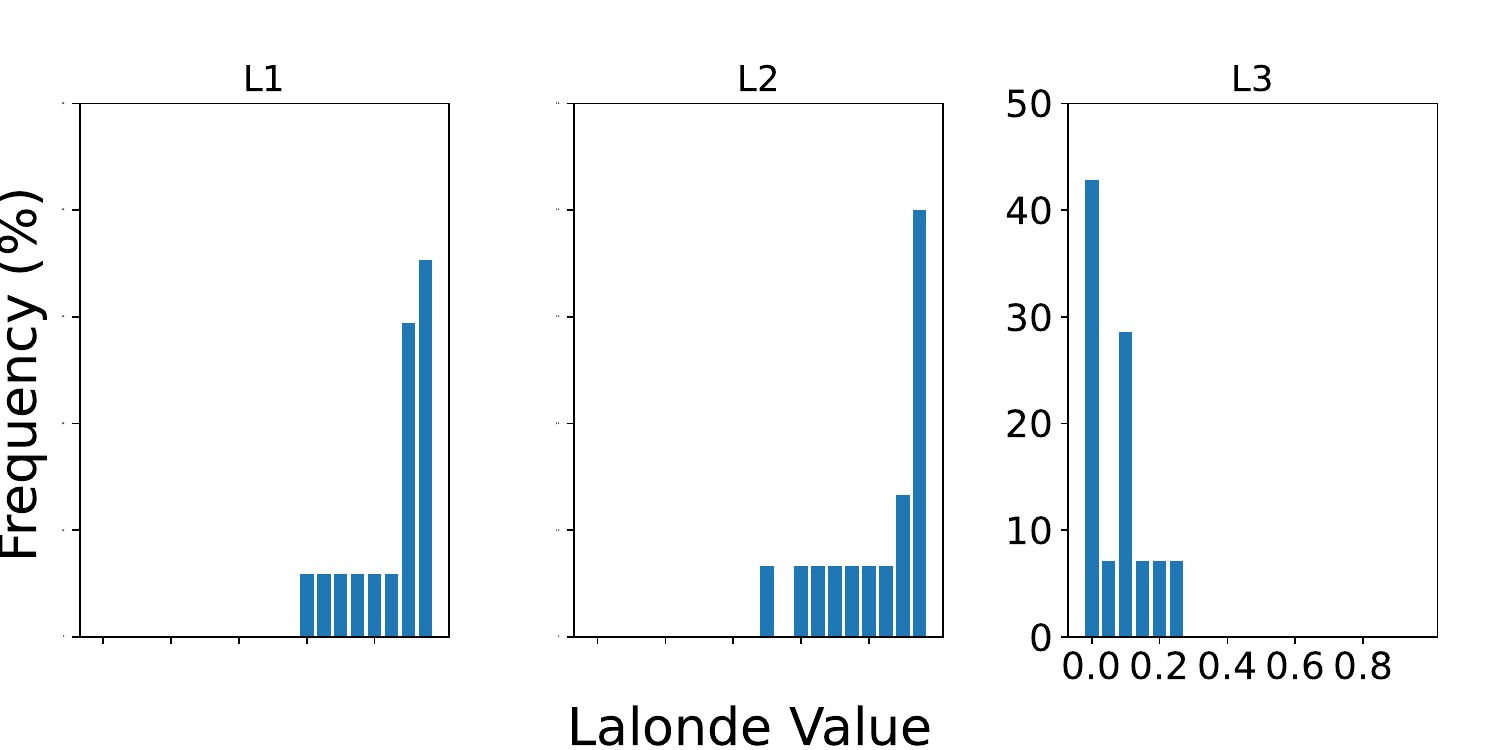}
        \caption{Cyclist side to radar.}
        \label{fig: Cyclist0-10}
    \end{subfigure}
    \begin{subfigure}[b]{0.3\textwidth}
        \includegraphics[width=\textwidth]{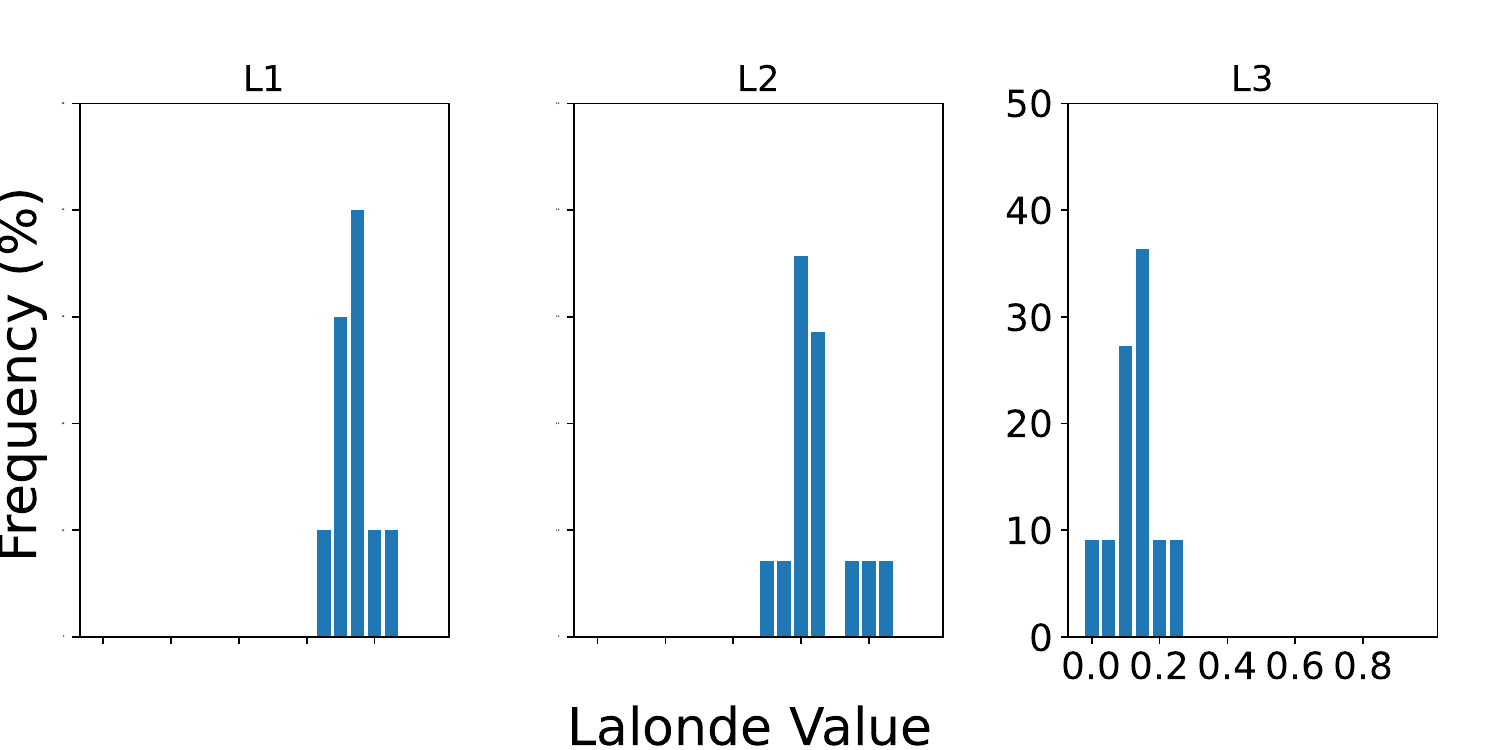}
        \caption{Cyclist skew to radar.}
        \label{fig: Cyclist40-50}
    \end{subfigure}
    \begin{subfigure}[b]{0.3\textwidth}
        \includegraphics[width=\textwidth]{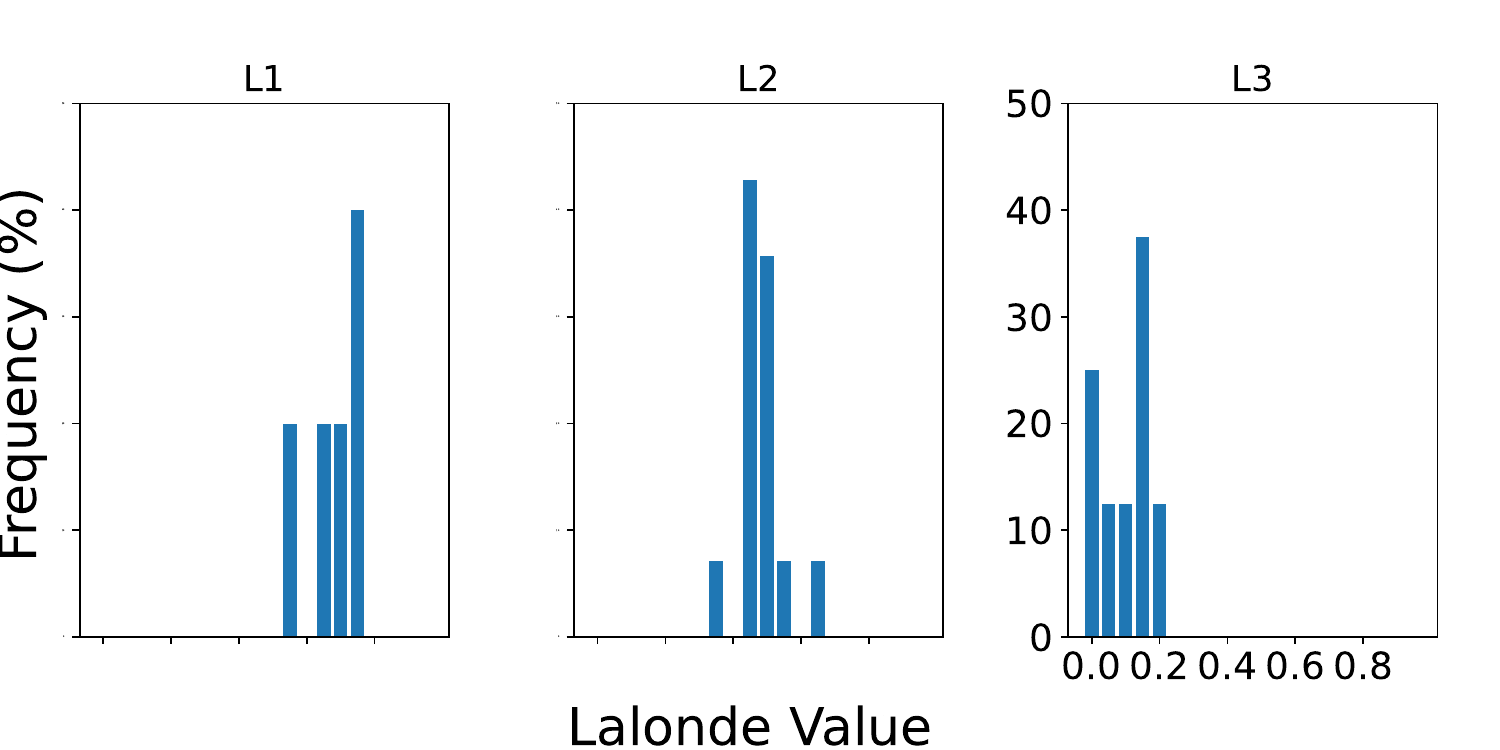}
        \caption{Cyclist face to radar.}
        \label{fig: Cyclist80-90}
    \end{subfigure}
    \begin{subfigure}[b]{0.3\textwidth}
        \includegraphics[width=\textwidth]{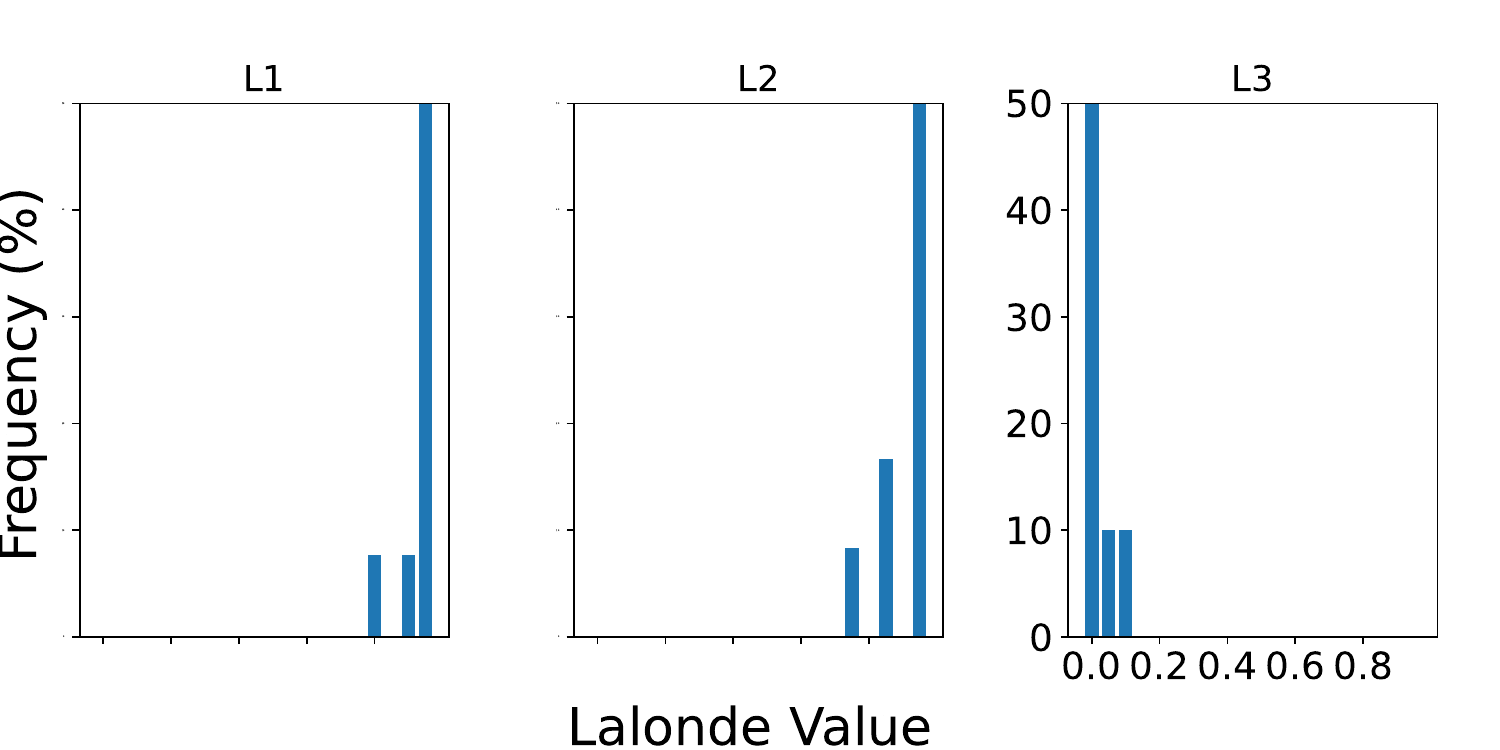}
        \caption{Pedes. side to radar.}
        \label{fig: Pedestrian0-10}
    \end{subfigure}
    \begin{subfigure}[b]{0.3\textwidth}
        \includegraphics[width=\textwidth]{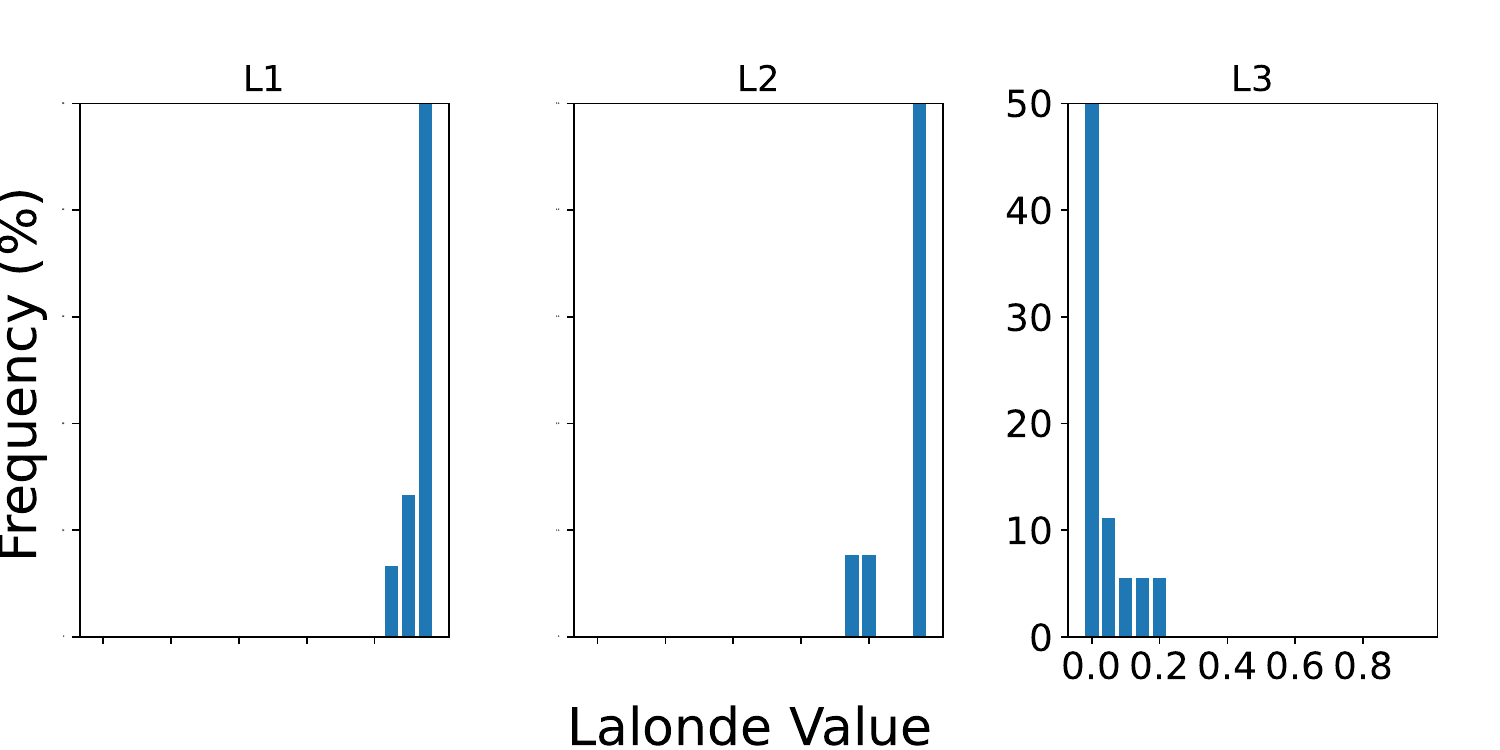}
        \caption{Pedes. side to radar.}
        \label{fig: Pedestrian40-50}
    \end{subfigure}
    \begin{subfigure}[b]{0.3\textwidth}
        \includegraphics[width=\textwidth]{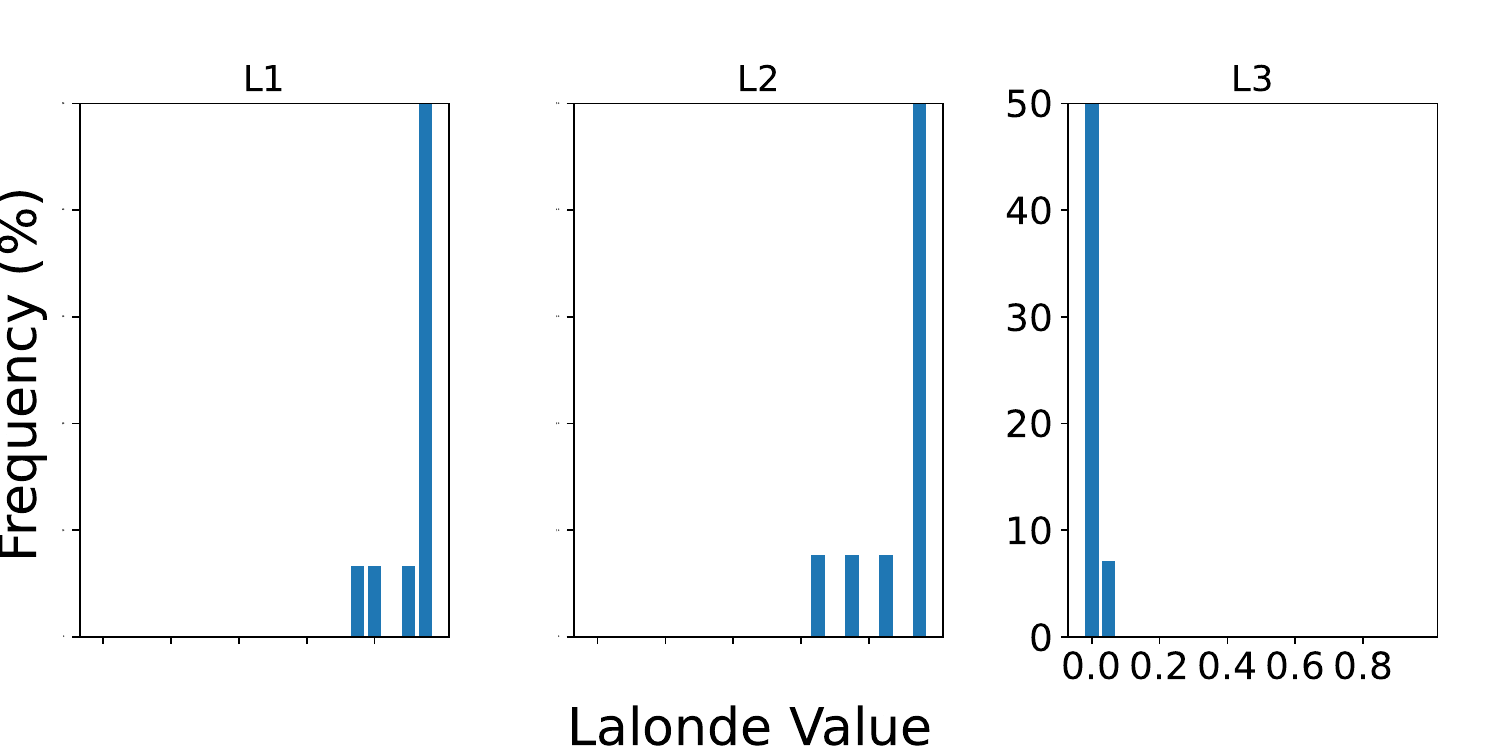}
        \caption{Pedes. side to radar.}
        \label{fig: Pedestrian80-90}
    \end{subfigure}
    \caption{Lalonde features for car, cyclist, and pedestrian in different orientations.}
    \label{fig: Lalonde_Feature}
\vspace{-6mm}
\end{figure}

The Lalonde features \(L_1\), \(L_2\), and \(L_3\) for cars, cyclists, and pedestrians across different orientations in the VoD dataset are depicted in Fig. \ref{fig: Lalonde_Feature}. Here, cars exhibit more pronounced planarity features in $L_1$ and $L_2$ compared to other objects, while cyclists can be more effectively detected from the side due to prominent point and planar features. Pedestrians consistently exhibit similar geometric characteristics across different orientations.

\begin{figure}[h]
    \vspace{-7mm}
    \centering
    \includegraphics[width=0.95\linewidth]{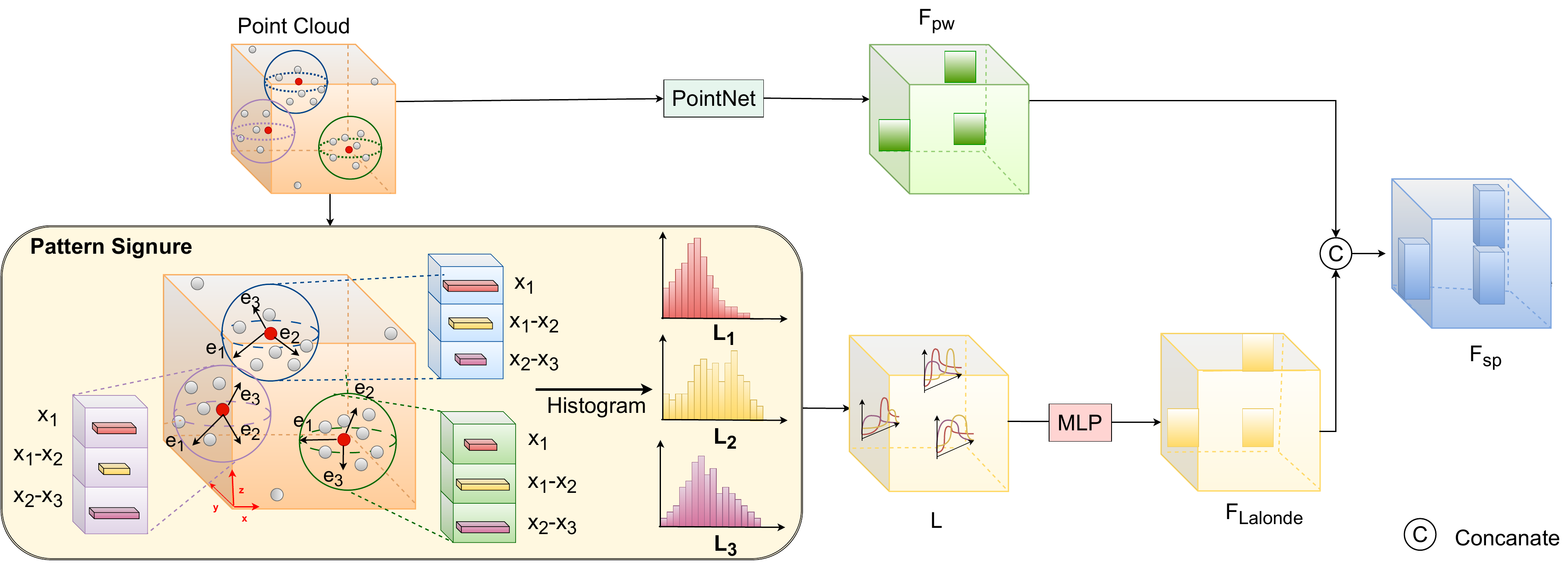}
    \caption{GeoSPA implementation.}
    \label{fig: GeoSPA}
\vspace{-8mm}  
\end{figure}

\paragraph{Implementation of GeoSPA:} We introduced GeoSPA based on Lalonde features to capture the detailed spatial pattern. 
Its structure is shown in Fig. \ref{fig: GeoSPA}.

On one hand, a PointNet \cite{qi2017pointnet} architecture is used to extract point-wised feature as Equation \ref{eq: F_pw}
\vspace{-2mm} 
\begin{equation}
F_{pw} = \text{PointNet}(S) = \max(\text{MLP}(S)),
\label{eq: F_pw}
\vspace{-2mm} 
\end{equation}
where \(F_{\text{pw}}\) is the point-wised features within the subset \(S\) through the application of MLP layers and max pooling in PointNet. 
On the other hand, for each point \(p_i \in S\), we compute the Lalonde descriptors \(L(p_i)\) within its fixed neighborhood \(N(p_i)\) and transform it into high-dimensional Lalonde features \(F_{\text{Lalonde}}(p_i)\) using an MLP layer. This process is captured by Equation \ref{eq: combined_F_L}
\vspace{-2mm}
\begin{equation}
F_{\text{Lalonde}}(p_i) = \text{MLP}(L(N(p_i))).
\label{eq: combined_F_L}
\vspace{-1mm}
\end{equation}
The final spatial feature, \(F_{\text{sp}}(p_i)\), is calculated by concatenating the point-wise features and the high-dimensional Lalonde features as Equation \ref{eq: F_C}
\vspace{-2mm}
\begin{equation}
\vspace{-2mm} 
F_{\text{sp}}(p_i) = F_{\text{pw}}(p_i) 
 \textcircled{c}  F_{\text{Lalonde}}(p_i),
\label{eq: F_C}
\vspace{-1mm} 
\end{equation}
where $\textcircled{c}$ denotes the concatenation operation.

We designed GeoSPA with plug-and-play functionality, allowing it to be embedded into any point-wise feature extraction process. In single-stage networks, GeoSPA is incorporated into the initial feature extraction branch. For two-stage networks with proposal refinement, GeoSPA can also operate within the Region of Interest (ROI) to enhance local spatial features except in a point-wise feature extraction branch, thereby aiding in the refinement of proposals.

\subsection{Distributed External Multi-View Attention}
Our DEMVA module leverages dual-projection techniques: cylindrical and BEV views with external attention mechanisms. The multi-view extraction captures global features from multiple perspectives, while the distributed attention mechanism incrementally learns shared information across the entire dataset.

For BEV projection, each point \(p_n (x_n, y_n, z_n)\) is systematically allocated to a defined pillar as defined in Equation \ref{eq: BEV point to pillar}
\vspace{-2mm}
\begin{equation}
    \mathcal{M}_{BEV}(p^n_{BEV}) \rightarrow q^m_{BEV},
    \label{eq: BEV point to pillar}
\vspace{-2mm}
\end{equation}
where \(\mathcal{M}_{BEV}\) symbolizes the mapping operation that assigns points to pillars in the BEV projection, \(p^n_{BEV}\) denotes the $n^{th}$ point in BEV coordinates, and \(q^m_{BEV}\) denotes the $m^{th}$ pillar.
In the cylindrical projection, a transformation of points into cylindrical coordinates \((\rho_n, \theta_n, z'_n)\) is applied to maintain the fidelity of vertical dimensions as below:
\vspace{-3mm}
\begin{equation*}
\rho_n = \sqrt{x_n^2 + y_n^2}, \quad \theta_n = \text{arctan}\left(\frac{y_n}{x_n}\right), \quad z'_n = z_n,
\label{eq: transform}
\vspace{-3mm}
\end{equation*}
where \(\rho_n\) is the radial distance from the origin to the $n^{th}$ point, \(\theta_n\) specifies the angular orientation around the z-axis, and \(z'_n\) upholds the original elevation data. Following this transformation, a similar pillar structuring is implemented in the cylindrical space as Equation \ref{eq: modified cylinder point to pillar}
\vspace{-2mm}
\begin{equation}
    \mathcal{M}_{cyl}(p^n_{cyl}) \rightarrow q^m_{cyl},
    \label{eq: modified cylinder point to pillar}
\vspace{-2mm}
\end{equation}
where \(\mathcal{M}_{cyl}\) means the function that maps the $n^{th}$ point to the $m^{th}$ pillar. 

Later, \(q^m_{BEV}\) and \(q^m_{cyl}\) are mapped onto pseudo images. Utilizing two 2D CNNs, we further extract global features \(I_{BEV}\) and \(I_{cyl}\) with dimension \((B, C \times Z, H, W)\) from the BEV and cylindrical pseudo images. Here, \(B\) denotes the batch size, \(C\) indicates the number of channels, \(Z\) is the feature depth, \(H\) and \(W\) refer to the height and width of the feature maps.

Strong dataset-wide correlations, such as vehicles predominantly occupying central positions while pedestrians and cyclists are often at the edges, help to enhance the understanding of the road environment. Furthermore, objects within the same category always exhibit similar shapes and motion patterns. Therefore, we employ external attention mechanisms, effectively capturing and encoding shared information across the entire dataset.

\begin{figure}[h]
    \vspace{-3mm}
    \centering
    \includegraphics[width=0.9\linewidth]{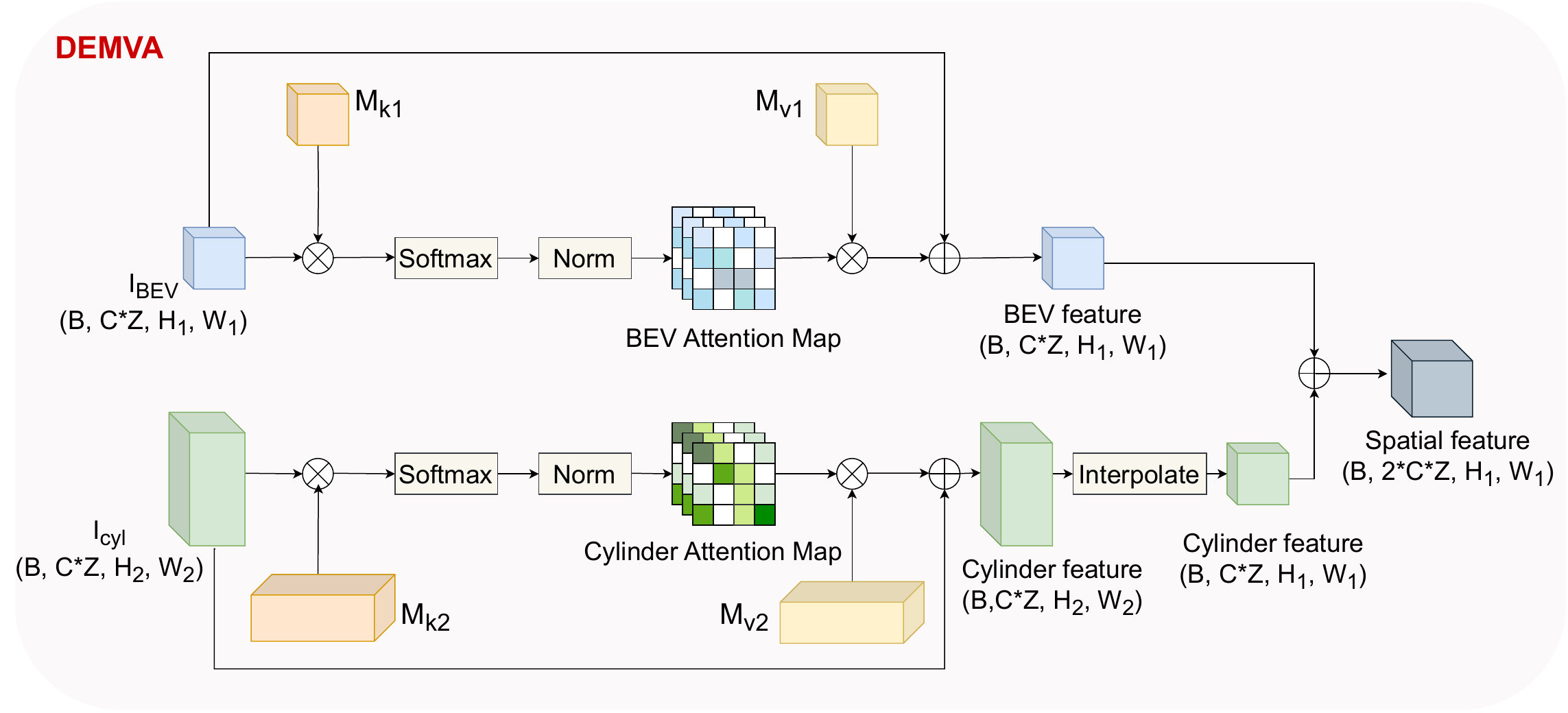}
    \caption{Distributed Multi-View Spatial Attention}
    \label{fig: external attention}
    \vspace{-8mm}
\end{figure}

Fig. \ref{fig: external attention} shows the process of extracting dataset-wide features by DEMVA. Initially, the inputs undergo a preliminary transformation by a matrix \(M_k\), leading to an attention map via a Softmax and Normalization layer. This attention map is then multiplied with a second external matrix, \(M_v\), leading to the feature graph. The resultant feature graph is subsequently fused with the original cylindrical and BEV features.
The \(M_k\) and \(M_v\) are the key and value components in the attention mechanism. 
Both matrices, as the weight of one-dimensional convolution, are dynamically updated throughout the learning phase.
Therefore, the matrices effectively function as memory units, continuously integrating information from the new frame into the matrices, thus enabling the extraction of dataset-wide features.
As a result, the DEMVA applies this dataset-wide knowledge to each individual frame and facilitates the fusion of dataset-wide features with single-frame global features.

Given the inherent sparsity of radar point clouds, our DEMVA plays a crucial role in enhancing the model's performance by extracting global information through multiple perspectives and further enriching it with dataset-wide features via external attention mechanisms.

\section{Experiment}

\subsection{Dataset and Metrics}
The MUFASA network is evaluated on the VoD \footnote{https://github.com/tudelft-iv/view-of-delft-dataset} and TJ4DRadSet \footnote{https://github.com/TJRadarLab/TJ4DRadSet} dataset to validate its efficacy. The VoD dataset comprises 8,693 frames of synchronized and calibrated data from LiDAR, cameras, and 4D radar. The data is mainly acquired in busy urban environments, including diverse traffic components. We evaluated our methods on the validation set, as the test server is unavailable. The TJ4DRadSet dataset includes 40,000 frames of synchronized LiDAR, camera, and radar data, with 7,757 frames annotated. This dataset spans various scenarios, including elevated roads, complex crossroads,  single-direction lanes, and urban streets.

To assess our method, we employed Average Precision (AP) for each class and calculated the mean Average Precision (mAP) across all the classes. To calculate AP, we determined the Intersection over Union (IoU) between the ground truth and predicted 3D bounding boxes. A minimum IoU threshold of 50\% was set for the car and truck categories, while a threshold of 25\% was applied for pedestrian and cyclist categories. 
In the VoD dataset, the detection performance was analyzed in two distinct regions: the entire scene and the driving corridor, aligned with the original paper \cite{palffy2022multi}. 
For the TJ4DRadSet dataset, evaluations are conducted in both BEV and 3D perspectives as in \cite{zheng2022tj4dradset}.

\subsection{Implementation Details}

The models are trained for 80 epochs with a batch size of 8, utilizing 4 NVIDIA Tesla P40 GPUs. For training, we used the Adam optimizer \cite{kingma2014adam}, with an initial learning rate of 0.03 and a weight decay rate of 0.01.
Our implementation is based on OpenPCDet \cite{od2020openpcdet}, a comprehensive library designed for point cloud tasks. 
Data augmentation is implemented, including rotation, flipping, and scaling, to improve the models' robustness and generalization capabilities.

\subsection{Quantitaive Results}

We performed a comprehensive comparison of our proposed method with existing point cloud object detection techniques on the VoD and TJ4DRaDSet datasets. The results are detailed in Table \ref{tab: main result vod} and \ref{tab: main result TJ}.

\setlength\tabcolsep{1.9pt}
\begin{table}
\vspace{-6mm}
\centering
\caption{Comparative AP results on VoD val. set. The values are in \%. The best results are bold and the second best are marked with an underline.}\label{tab: main result vod}
\begin{tabular}{c|c|cccc|cccc}
\hline
\multirow{2}{*}{Methods} &\multirow{2}{*}{Sensor} & \multicolumn{4}{c|}{All area}   & \multicolumn{4}{c}{Driving Corridor}\\
\cline{3-10}
 &  &Car & Ped. & Cyc. & mAP & Car & Ped. & Cyc. & mAP\\
\hline
SECOND\textsuperscript{\dag}\cite{yan2018second} &R &32.35  &24.49  &51.44  &36.10  &67.98  &35.45  &72.30  &59.18 \\
RPFA-Net\textsuperscript{\dag}\cite{xu2021rpfa} &R &33.45 &26.42 &56.34 &38.75 &68.68 &34.25  &80.36  &62.44 \\
CenterPoint\cite{yin2021center} &R &33.87 &\underline{39.01} &66.85 &46.58 &62.98 &\underline{49.22} &85.35 &65.85 \\
PointPillars\cite{lang2019pointpillars} &R &37.92   &31.24  &65.66  &44.94  &71.41  &42.27  &87.68  &67.12 \\
MVFAN\textsuperscript{\dag}\cite{yan2023mvfan} &R &34.05 &27.27  &57.14  &39.42  &69.81  &38.65  &84.87  &64.38 \\
MVFAN\cite{yan2023mvfan} &R &38.12 &30.96  &66.17  &45.08  &71.45  &40.21  &86.63  &66.10 \\
PV-RCNN\cite{shi2020pv} &R &41.65   &38.82  &58.36  &46.28  &\underline{72.00}  &43.53  &78.32  &64.62 \\
\hline
BEVFusion\cite{liu2023bevfusion} &R+C &37.85 &\textbf{40.96}  &\textbf{68.95} &49.25 &70.21 &45.86 &\textbf{89.48} &68.52 \\
RCFusion\cite{zheng2023rcfusion} &R+C &\underline{41.70} &38.95  &68.31  &\underline{49.65}  &71.87  &47.50  &88.33  &69.23 \\
\hline
MUFASA(pp) &R &41.07  &37.52  &68.07  &48.89  &71.89  &47.40  &\underline{89.02}  &\underline{69.44}\\
MUFASA(pv) &R &\textbf{43.10}  &38.97  &\underline{68.65}  &\textbf{50.24}  &\textbf{72.50}  &\textbf{50.28}  &88.51  &\textbf{70.43}\\
\hline
\end{tabular}
\\[1pt]
\scriptsize{R denotes radar sensor, R+C denotes the fusion of radar and camera. The pp denotes MUFASA with Pointpillars head, and pv denotes MUFASA with PV-RCNN head. The results with \textsuperscript{\dag} are inherited from \cite{yan2023mvfan}. The results of BEVFusion and RCFusion are reported from \cite{xiong2023lxl}.}
\vspace{-0mm}
\end{table}

Table \ref{tab: main result vod} demonstrates the remarkable performance of our method, surpassing state-of-the-art radar-based object detection methods on the VoD dataset.
Leveraging the flexibility of key modules, the MUFASA can be embedded into different detection heads. The results also show a consistent enhancement in performance across different detection architectures. 
Notably, when incorporated with the PV-RCNN detection head, our approach achieves the highest mAP of 50.24\% over the entire area and 70.43\% within the driving corridor. This marks a significant improvement of 3.96\% and 5.81\%, respectively, over the PV-RCNN performance metrics. 

Furthermore, compared to radar and camera fusion methodologies such as BEVFusion \cite{liu2023bevfusion} and RCFusion \cite{zheng2023rcfusion}, our approach maintains competitive performance with the highest mAP, underscoring its efficacy and robustness in capturing comprehensive features.

\setlength\tabcolsep{1.0pt}
\begin{table}
\vspace{-7mm}
\centering
\caption{Comparative AP results on TJ4DRaDSet test set. The values are in \%. The best results are bold and the second best are marked with an underline.}\label{tab: main result TJ}
\begin{tabular}{c|c|ccccc|ccccc}
\hline
\multirow{2}{*}{Methods} &\multirow{2}{*}{Sensor} & \multicolumn{5}{|c|}{3D}   & \multicolumn{5}{c}{BEV}\\
\cline{3-12}
 & &  Car & Ped. & Cyc. & Truck & mAP & Car & Ped. & Cyc. & Truck & mAP\\
\hline
PV-RCNN\cite{shi2020pv}&R &19.46 &9.34 &40.44 &16.09 &21.33 &29.94 &9.73 &45.12 &22.92 &26.93\\
PartA$^2$\textsuperscript{\dag}\cite{shi2019part} &R&18.65 &23.28 &44.14 &9.63 &23.92 &29.95 &24.31 &49.08 &15.05 &29.60\\
SECOND\textsuperscript{\dag}\cite{yan2018second} &R &18.18& 24.43 &32.36 &14.76 &22.43 &36.02 &28.58 &39.75 &19.35 &30.93\\
PointPillars\cite{lang2019pointpillars}&R &23.98 &18.70 &43.25 &17.23 &25.79 &36.20 &20.47 &46.81 &\underline{30.60} &33.52\\
MVFAN\cite{yan2023mvfan} &R &23.37 &23.58 &45.35 &17.62 &27.48 &33.79 &25.92 &47.96 &30.15 &34.46\\
RPFA-Net\textsuperscript{\dag}\cite{xu2021rpfa} &R &26.89&\textbf{27.36}&\underline{50.95}&14.46&29.91&\textbf{42.89}&\underline{29.81}&\underline{57.09}&25.98&38.94\\
\hline
MVX-Net\textsuperscript{\dag}\cite{sindagi2019mvx} &R+C&22.28 &19.57 &50.70 &11.21 &25.94 &37.46 &22.70 &54.69 &18.07 &33.23\\
RCFusion\textsuperscript{\dag}\cite{zheng2023rcfusion} &R+C &\textbf{29.72} &\underline{27.17} &\textbf{54.93}& \underline{23.56} &\textbf{33.85} &40.89&\textbf{30.95} &\textbf{58.30} &28.92&\textbf{39.76}\\
\hline
MUFASA(pp)&R &\underline{27.86} &25.42 &42.58 &19.61 &28.87 &\underline{41.72} &27.93 &46.42 &28.69 &36.19\\
MUFASA(pv) &R& 23.56&23.70 &48.39 &\textbf{25.25} &\underline{30.23} &41.25 &24.54 &53.64 &\textbf{36.97} &\underline{39.10}\\
\hline
\end{tabular}
\\[1pt]
\scriptsize{R denotes radar sensor, R+C denotes the fusion of radar and camera. The pp denotes MUFASA with Pointpillars head, and pv denotes MUFASA with PV-RCNN head. The results with \textsuperscript{\dag} are inherited from \cite{zheng2023rcfusion}.}
\vspace{-7mm}
\end{table}

From Table \ref{tab: main result TJ}, MUFASA demonstrates robust performance in radar-based object detection, achieving the highest mAP among radar-only methods, with 30.23\% for 3D detection and 39.10\% for BEV detection. Notably, MUFASA excels in detecting trucks with PV-RCNN head, highlighting its effectiveness in large object detection. Furthermore, MUFASA with the PV-RCNN head (pv) significantly outperforms the PointPillars head (pp) in detecting cyclists and trucks on the TJ4DRaDSet dataset, showing improvements of 5.81\% and 5.64\% in 3D detection, respectively. This enhanced performance is due to the GeoSPA module in the ROI stage, which is particularly effective at refining bounding boxes with an obvious length-width ratio.
However, in comparison to radar and camera fusion methods, such as MVX-Net \cite{sindagi2019mvx} and RCFusion \cite{zheng2023rcfusion}, MUFASA still exhibits a minor performance gap.

\subsection{Qualitative Results}
\begin{figure}[ht]
    \centering
    \vspace{-3mm}
        \begin{subfigure}{\textwidth}
            \vspace{-1mm}
            \centering
            \includegraphics[width=0.32\linewidth]{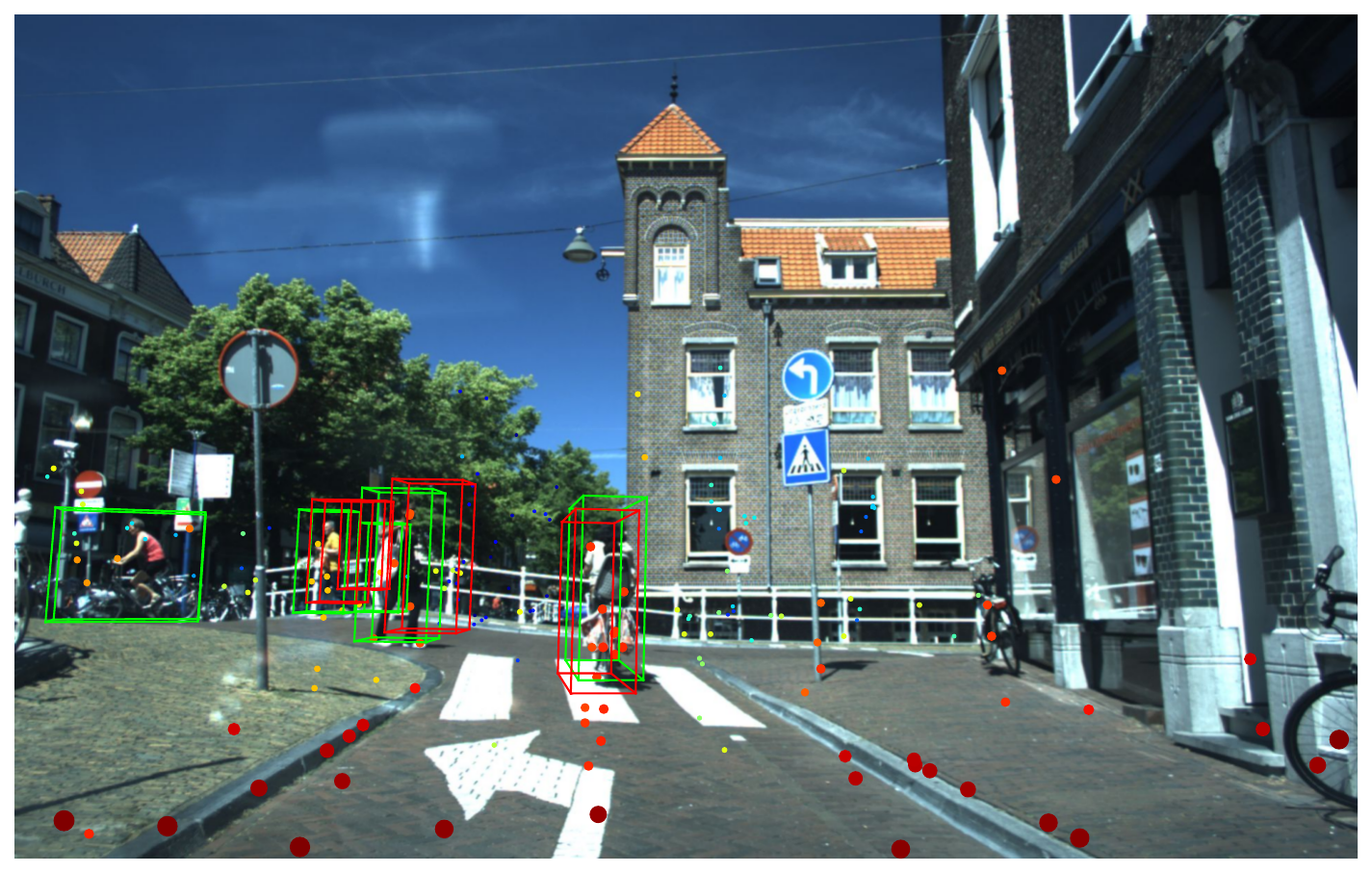}\hfill
            \includegraphics[width=0.32\linewidth]{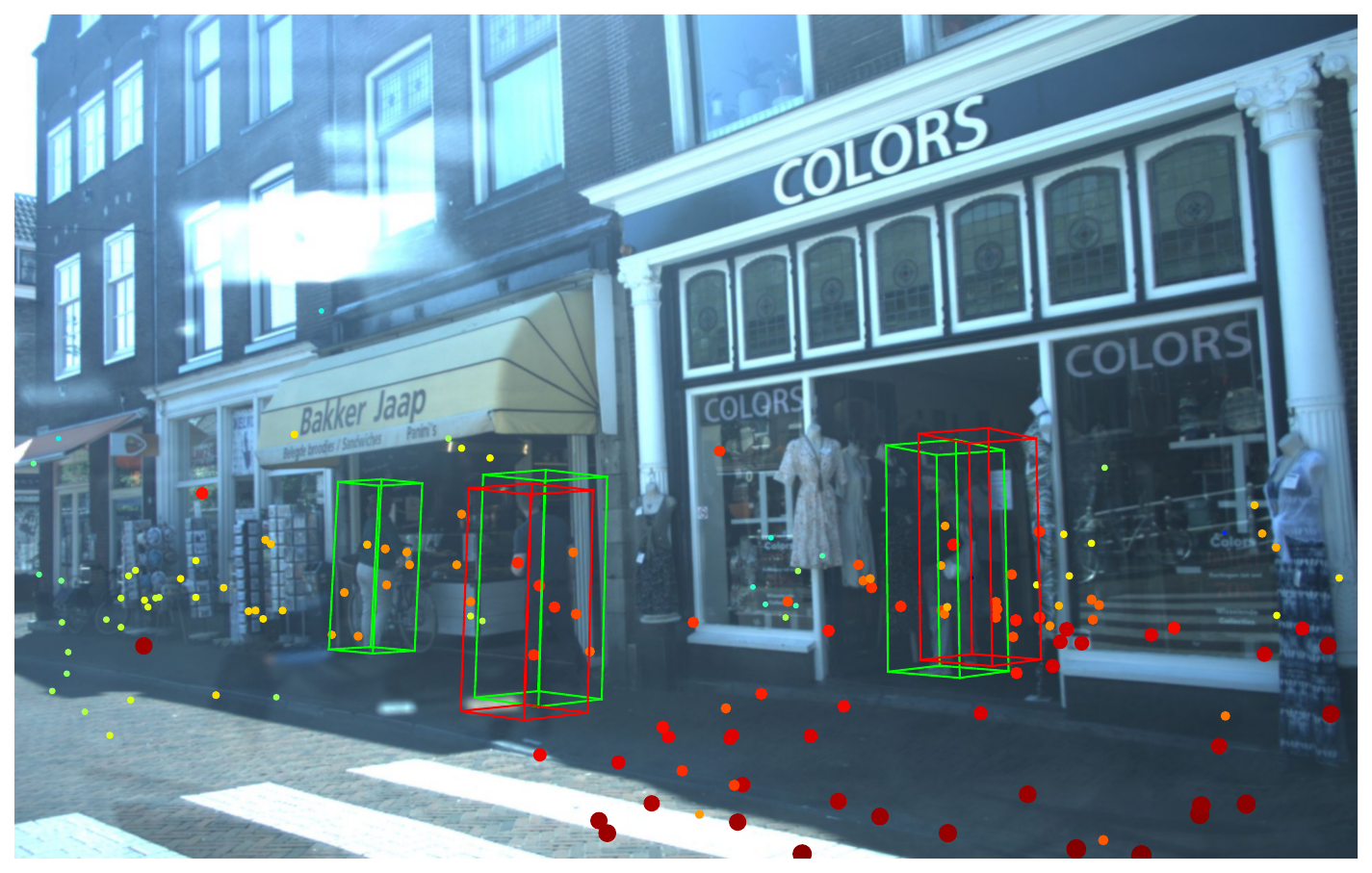}\hfill
            \includegraphics[width=0.32\linewidth]{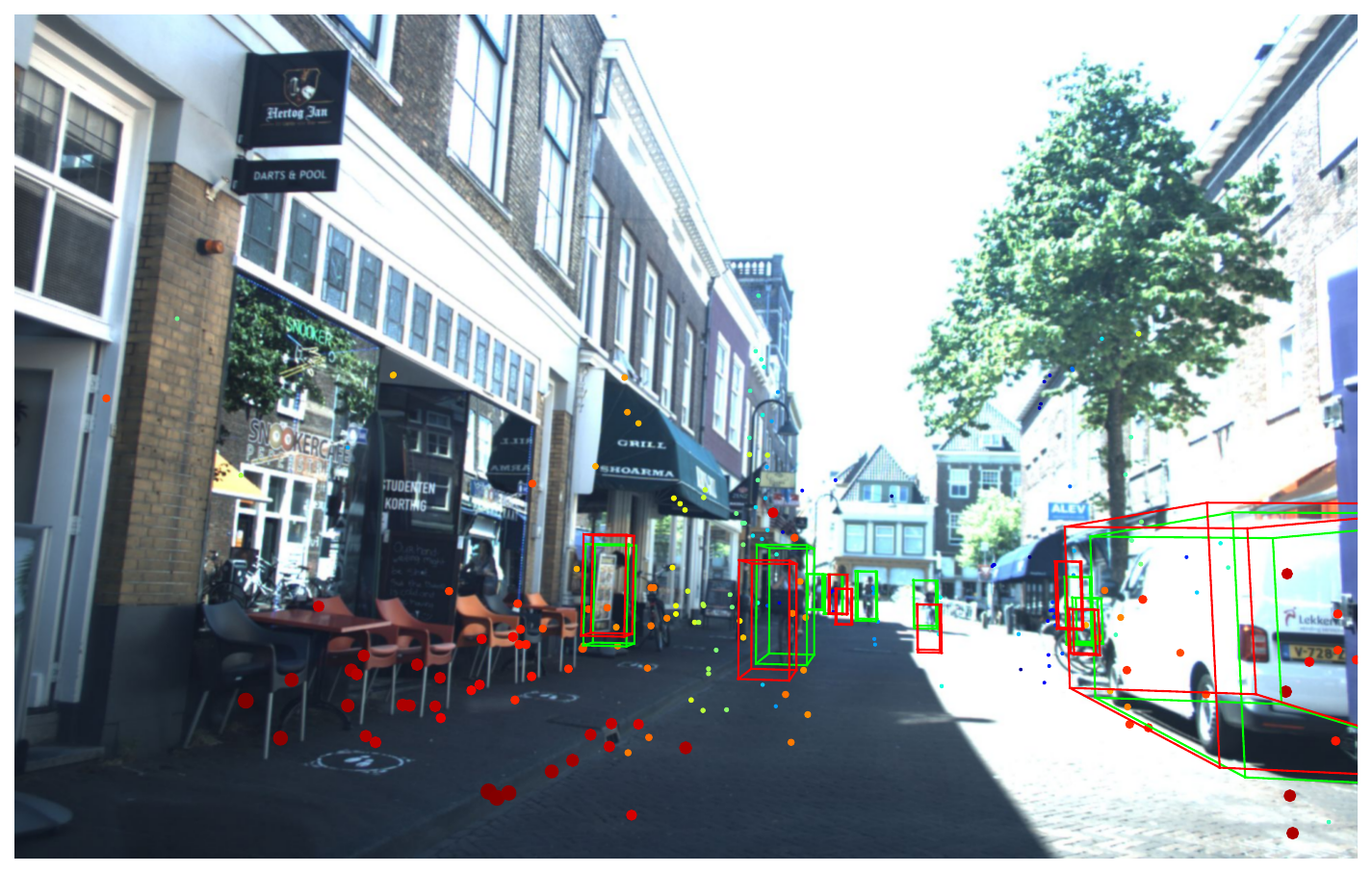}
            \vspace{-1mm}
            \caption{Results from PointPillars.}
            \label{fig:first_row}
            \vspace{-1mm}
        \end{subfigure}

        \begin{subfigure}{\textwidth}
            \centering
            \includegraphics[width=0.32\linewidth]{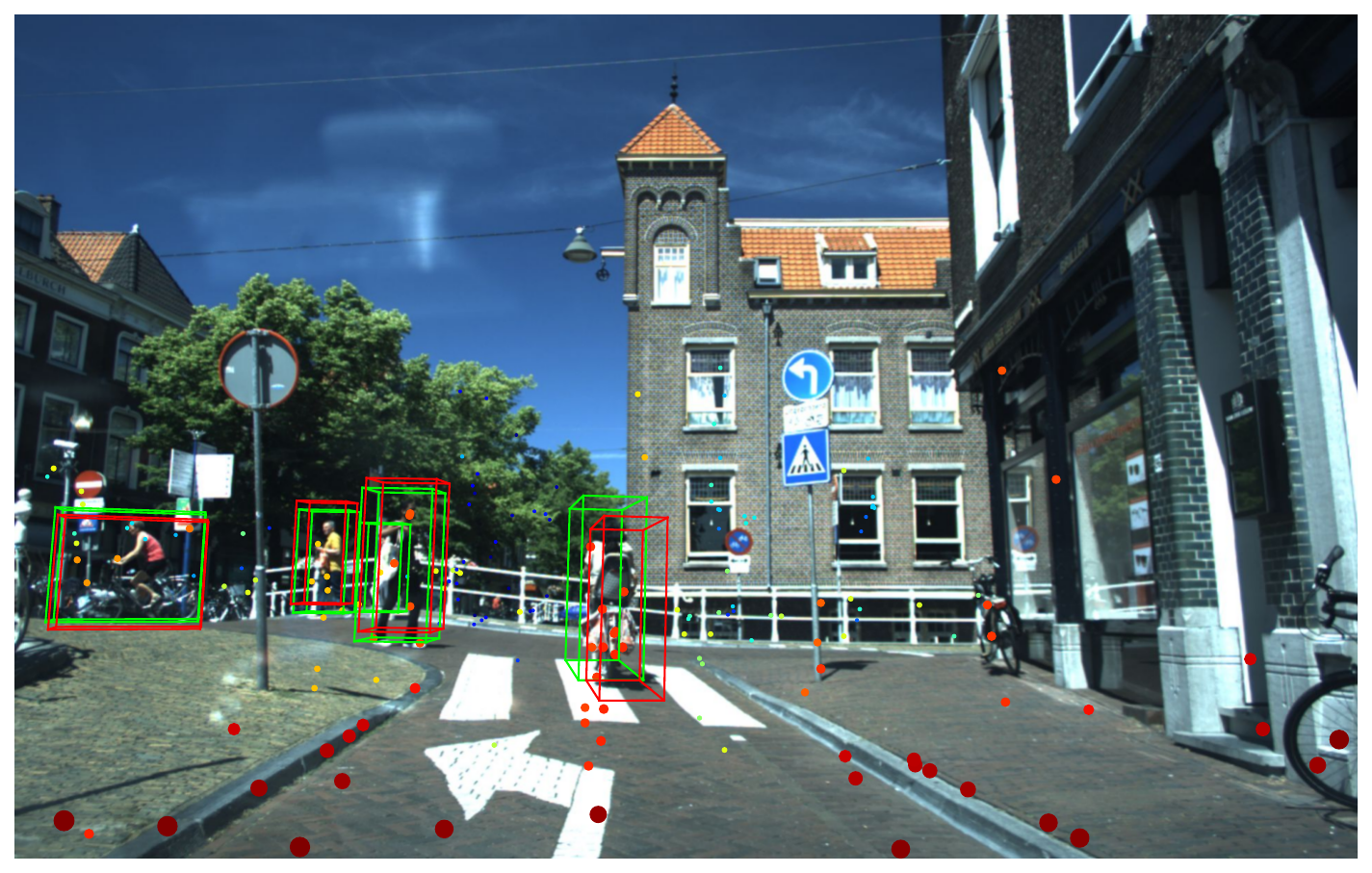}\hfill
            \includegraphics[width=0.32\linewidth]{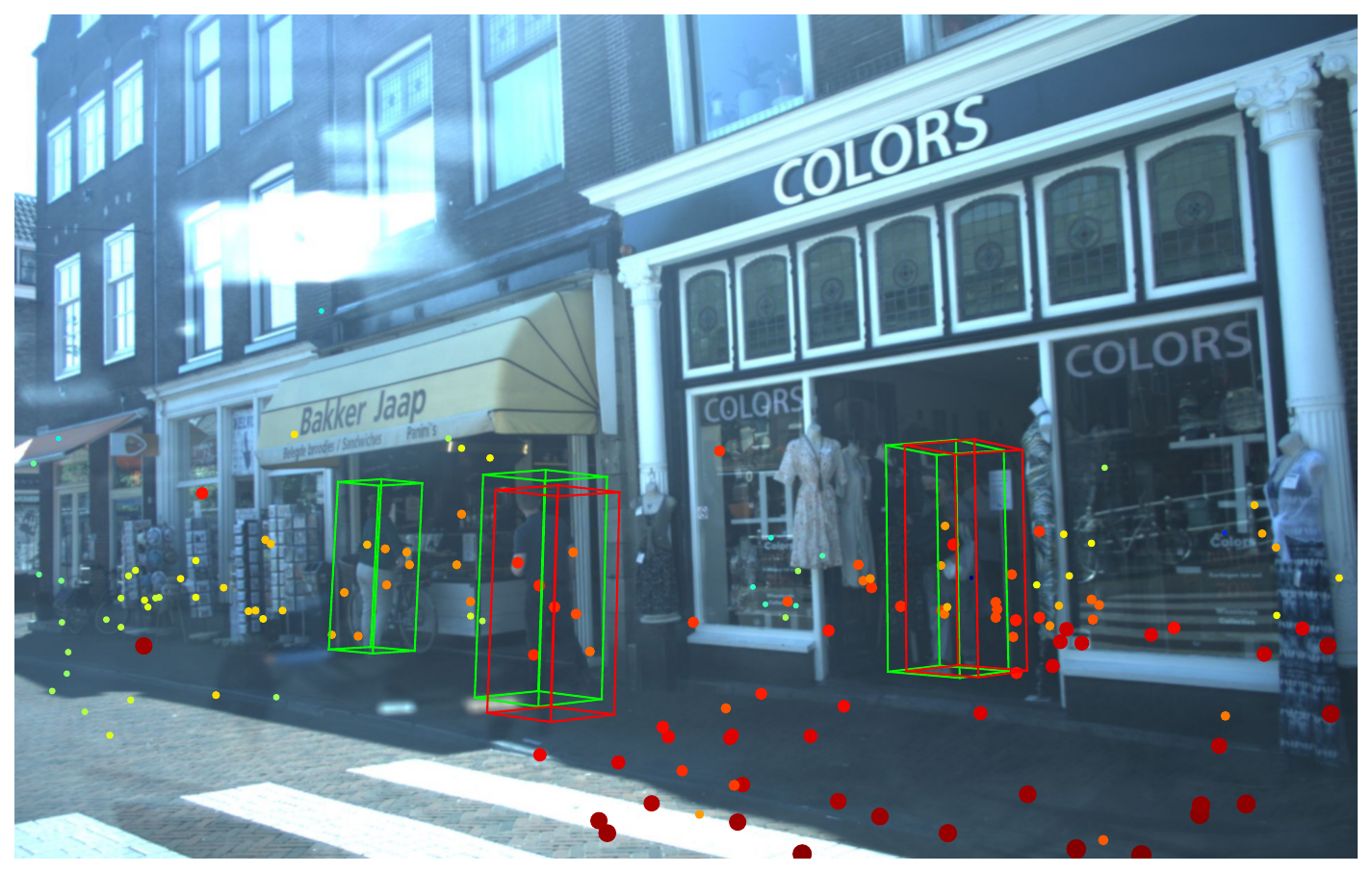}\hfill
            \includegraphics[width=0.32\linewidth]{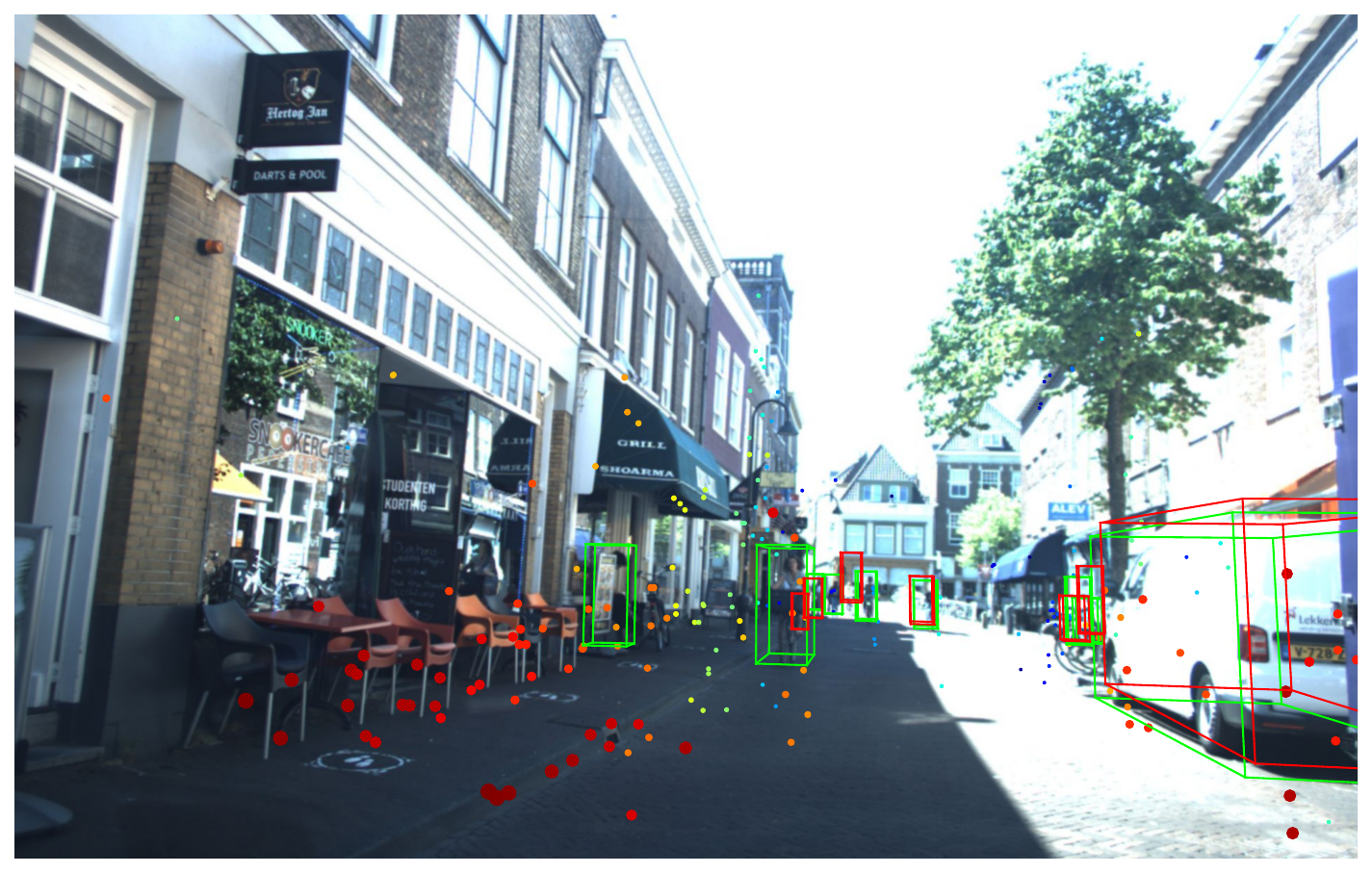}
            \vspace{-1mm}
            \caption{Results from PV-RCNN.}
            \label{fig:first_row}
            \vspace{-1mm}
        \end{subfigure}
        \begin{subfigure}{\textwidth}
            \centering
            \includegraphics[width=0.32\linewidth]{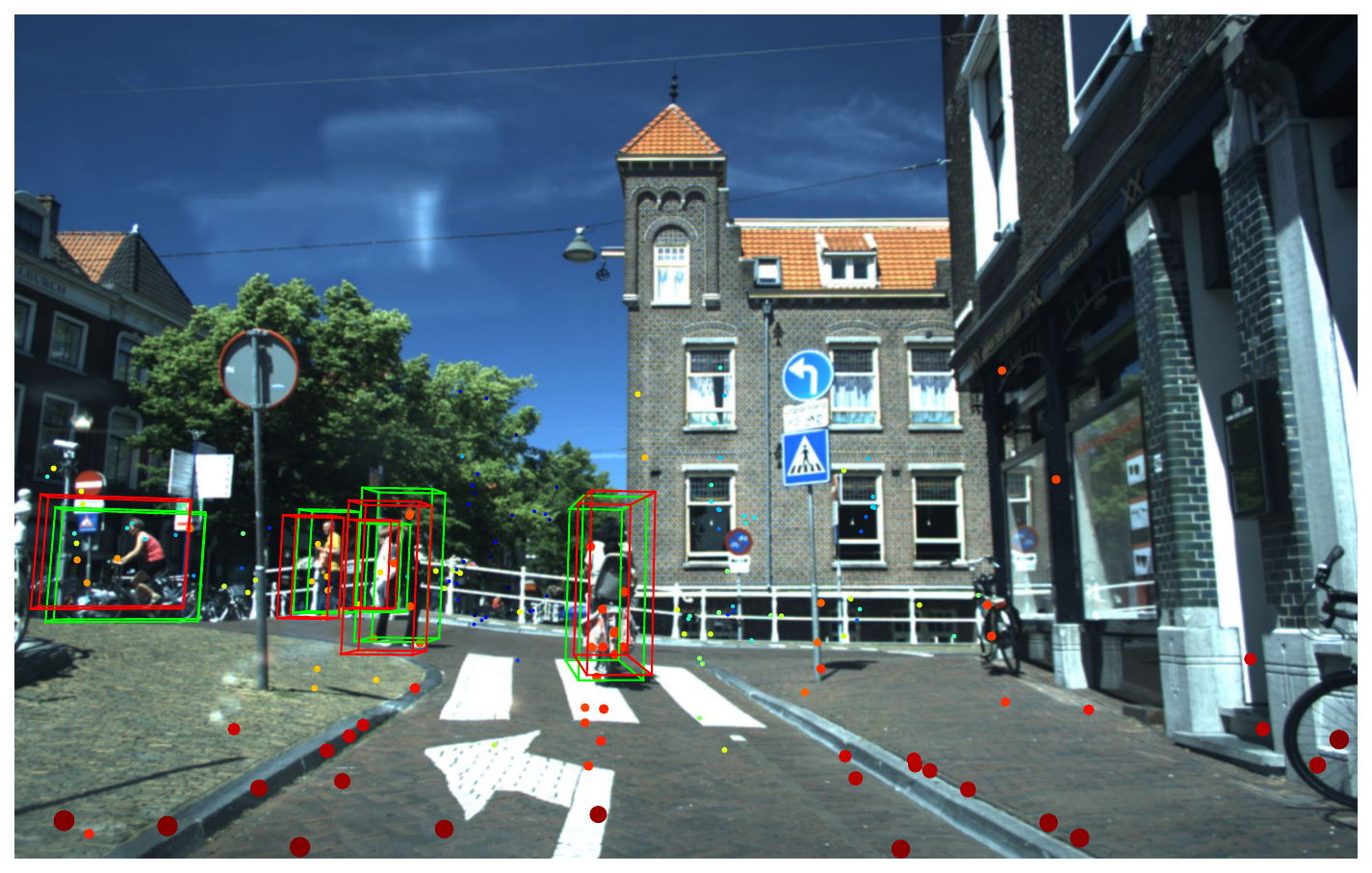}\hfill
            \includegraphics[width=0.32\linewidth]{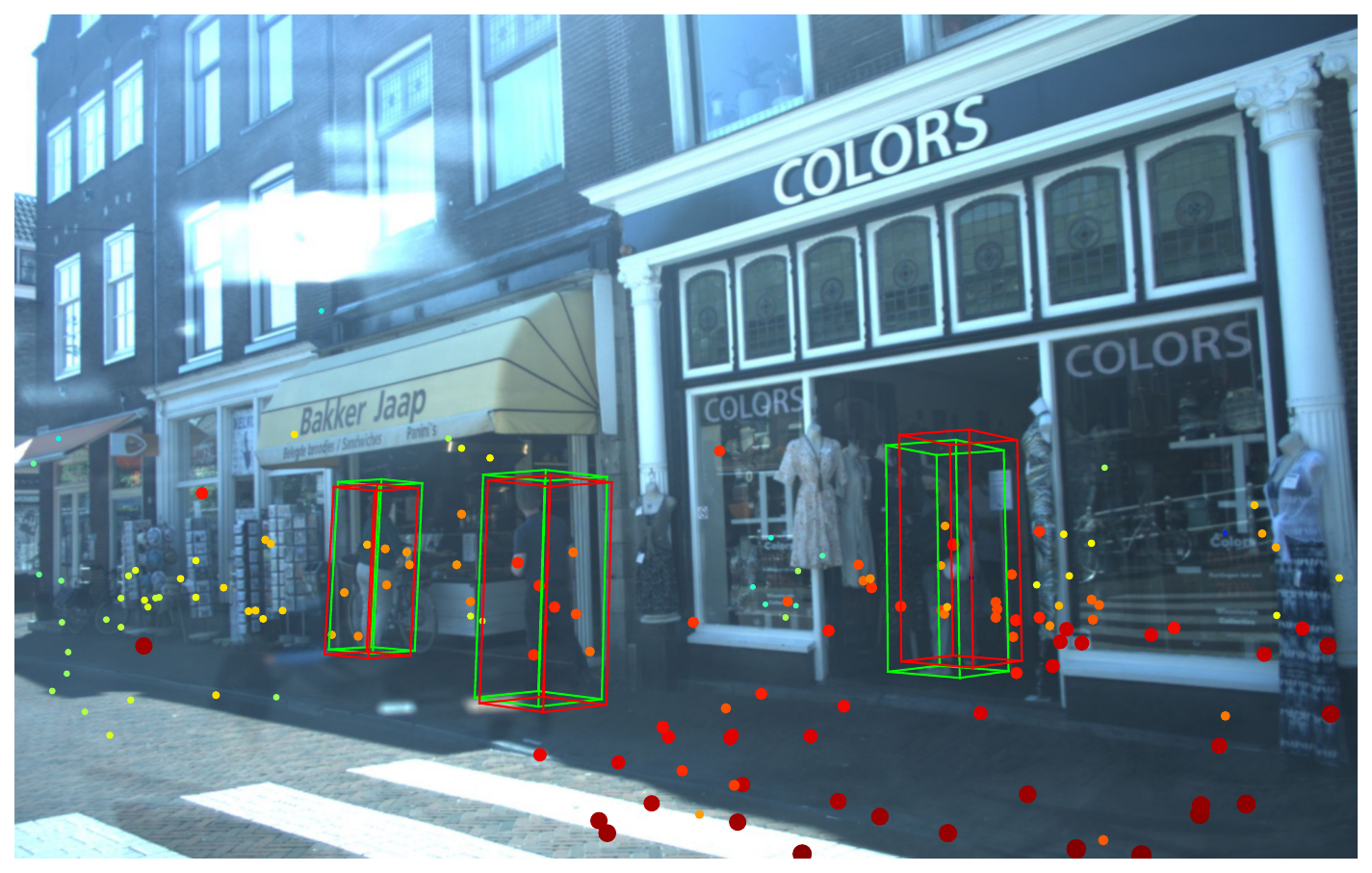}\hfill
            \includegraphics[width=0.32\linewidth]{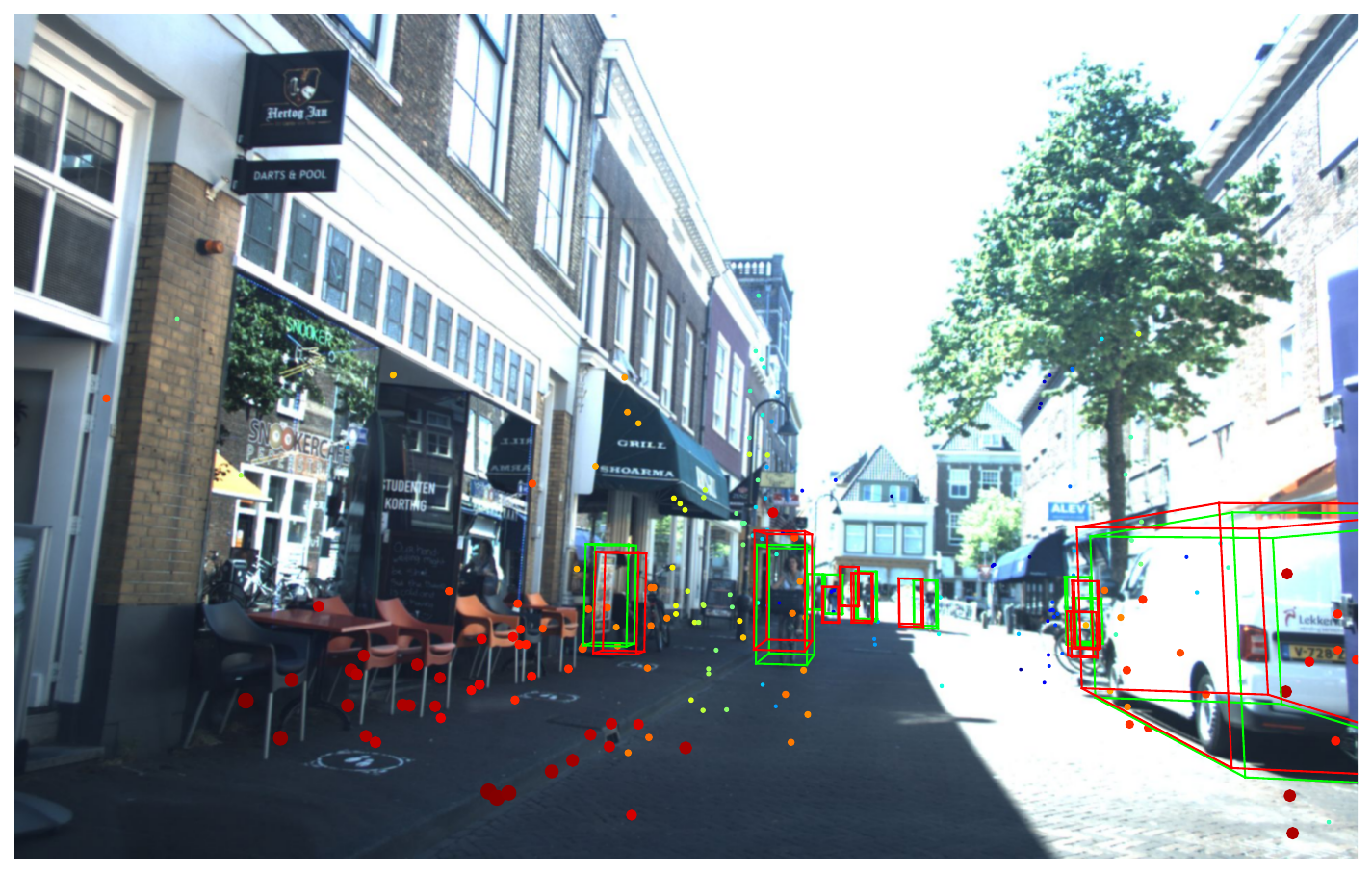}
            \vspace{-1mm}
            \caption{Results from MUFASA (Ours).}
            \label{fig:second_row}
            \vspace{-1mm}
        \end{subfigure}
    \vspace{-5mm}
    \caption{Visualization of detection.}
    \label{fig: overall}
    \vspace{-4mm}
\end{figure}

We visualize the bounding boxes generated by PointPillars\cite{lang2019pointpillars}, PV-RCNN\cite{shi2020pv}, and MUFASA on the VoD dataset. Ground truth boxes are marked in green, while predictions are in red. From Fig. \ref{fig: overall}, it is evident that MUFASA performs better detection with fewer false negative bounding boxes.

\subsection{Ablation Study}
The effectiveness of the critical modules are illustrated through an ablation study. Additionally, the generalization of our methods is demonstrated by incorporating key models with different settings.

\subsubsection{Analysis of key modules}
We first removed the DEMVA and GeoSPA from the MUFASA architecture. The remaining network was retrained with the same parameters as a baseline. Subsequently, we incrementally reintegrated each module into the baseline to observe their individual contributions. 
\setlength\tabcolsep{3pt}
\begin{table}
\vspace{-8mm}
\centering
\caption{Module analysis on VoD val. set. The values are in \%.}\label{tab: module analysis vod}
\begin{tabular}{cc|cccc|cccc}
\hline
 \multicolumn{2}{c|}{Module}& \multicolumn{4}{c|}{All area}   & \multicolumn{4}{c}{Driving Corridor}\\
\hline
D &G &  Car & Ped. & Cyc. & mAP & Car & Ped. & Cyc. & mAP\\
\hline
&  & 38.12 &30.96 &66.17 &45.08 &71.45 &40.21 &86.63 &66.10 \\
$\checkmark$  & &40.86 &32.11 &67.96 &46.98 &72.04 &42.94 &86.74 &67.24 \\
 &$\checkmark$ &41.25 &37.10 &66.95 &48.43 &71.86 &47.84 &87.08 &68.93 \\
$\checkmark$ &$\checkmark$ &\textbf{43.10}  &\textbf{38.97}  &\textbf{68.65}  &\textbf{50.24}  &\textbf{72.50}  &\textbf{50.28}  &\textbf{88.51}  &\textbf{70.43} \\
\hline
\end{tabular}
\\[1pt]
\scriptsize{D denotes the DEMVA module. G denotes the GeoSPA module. The detection heads are kept as PV-RCNN heads for better comparison.}
\label{tab: module analysis}
\vspace{-7mm}
\end{table}

The analysis from Table \ref{tab: module analysis} reveals that the DEMVA module enhances the detection performance for all the classes by integrating the dataset-wide features into each frame. Meanwhile, the GeoSPA module is particularly effective in improving the detection of pedestrians, which have fewer points. For instance, incorporating GeoSPA into our framework results in a substantial increase in pedestrian AP by 6.14\% and 7.63\% across all areas and the driving corridor, respectively, compared to the baseline. The main reason is our GeoSPA effectively captures the local features. Pedestrians, despite their varied movement patterns, maintain relatively consistent shapes. Learning the detailed geometric information in the neighborhood helps to identify the features that are uniquely from one category.

\subsubsection{Analysis of Different Settings}

To validate the rational layout of key modules, we strategically deployed the DEMVA and GeoSPA modules at different locations within the network architecture.
%\vspace{-2mm}
\paragraph{\textbf{GeoSPA on different locations:}}
Table \ref{tab: stage analysis vod} analyzes the impact of the GeoSPA module when applied at various locations within the network. Initially, we establish a baseline scenario where GeoSPA is not employed. Then, we apply GeoSPA in the early stage for the point-wise feature extraction and in the later stage to refine proposals with the PV-RCNN detection head.
\setlength\tabcolsep{3pt}
\begin{table}
\vspace{-8mm}
\centering
\caption{Layout of GeoSPA analysis on VoD val. set. The values are in \%.}\label{tab: module analysis vod}
\begin{tabular}{cc|cccc|cccc}
\hline
 \multicolumn{2}{c|}{Layout}& \multicolumn{4}{c|}{All area}   & \multicolumn{4}{c}{Driving Corridor}\\
\hline
1st-Stage &2nd-Stage &  Car & Ped. & Cyc. & mAP & Car & Ped. & Cyc. & mAP\\
\hline
 &  &40.86 &32.11 &67.96 &46.98 &72.04 &42.94 &86.74 &67.24 \\
$\checkmark$ & &42.52 &36.95 &\textbf{68.89} &49.45 &72.19 &48.84 &86.89 &69.31 \\
&$\checkmark$ &41.67 &33.94 &67.49 &47.70 &72.36 &46.20 &87.79 &68.78 \\
$\checkmark$ &$\checkmark$ &\textbf{43.10}  &\textbf{38.97}  &68.65  &\textbf{50.24}  &\textbf{72.50}  &\textbf{50.28}  &\textbf{88.51}  &\textbf{70.43} \\
\hline
\end{tabular}
\\[1pt]
\scriptsize{1st Stage denotes point-wised feature extraction branch. 2nd Stage denotes proposal refinement in ROI. The detection heads are kept as PV-RCNN heads for analysis.}
\vspace{-6mm}
\label{tab: stage analysis vod}
\end{table}

The outcomes indicate that integrating GeoSPA at both stages yields improvements in mAP for the entire area as well as the driving corridor.

%\vspace{-2mm}
\paragraph{\textbf{DEMVA on different branches:}}
We implement external attention to different projection branches. 
\setlength\tabcolsep{3pt}
\begin{table}
\vspace{-8mm}
\centering
\caption{Layout of External Attention (EA) analysis on VoD val. set. The values are in \%.}\label{tab: module analysis vod}
\begin{tabular}{cc|cccc|cccc}
\hline
 \multicolumn{2}{c|}{Layout Branch}& \multicolumn{4}{c|}{All area}   & \multicolumn{4}{c}{Driving Corridor}\\
\hline
BEV &Cylinder &  Car & Ped. & Cyc. & mAP & Car & Ped. & Cyc. & mAP\\
\hline
 &  &41.25 &37.10 &66.95 &48.43 &71.86 &47.84 &87.08 &68.93 \\
$\checkmark$ &  &41.98 &37.58 &68.13 &49.23 &71.89 &48.82 &87.56 &69.42 \\
&$\checkmark$ &42.55 &38.19 &68.18 &49.64 &72.11 &49.43 &88.14 &69.89 \\
$\checkmark$ &$\checkmark$ &\textbf{43.10}  &\textbf{38.97}  &\textbf{68.65}  &\textbf{50.24}  &\textbf{72.50}  &\textbf{50.28}  &\textbf{88.51}  &\textbf{70.43} \\
\hline
\end{tabular}
\\[1pt]
\scriptsize{The detection heads are kept as PV-RCNN heads for analysis.}
\vspace{-4mm}
\label{tab: branch analysis vod}
\end{table}
The results in Table \ref{tab: branch analysis vod} highlight the impact of external attention mechanisms within both the cylinder and BEV branches. Therefore, extracting hidden dataset-wide features with external attention mechanisms consistently enhances object detection accuracy. 

\section{Conclusion and Discussion}

Traditional radar object detection methods often struggle to fully exploit the detailed local and global features in a single frame, as well as the dataset-wide features across the entire dataset. 
In our method, MUFASA, the GeoSPA is designed as a flexible plug-and-play module, enhancing the model's capability to capture critical geometric details and recognize local spatial patterns across various detection networks.
Additionally, the DEMVA module synthesizes shared information across different frames and integrates it into the global features of each individual frame. 
Empirical evaluations on the VoD and TJ4DRaDSet datasets demonstrate the superiority of our approaches over existing radar point cloud-based object detection techniques. Specifically, our method achieved an mAP of 50.24\% and 70.43\% on the entire area and driving corridor of the VoD dataset, representing a notable improvement of 3.96\% and 5.81\%.

However, when compared to sensor fusion-based detection methods, our approaches may still exhibit a lower mAP. This observation underscores the need for future research focusing on integrating camera, LiDAR, and radar data. Such integration aims to improve detection accuracy and robustness under various environments, thereby expanding the practical use of our methods in real-world scenarios.

\section{Acknowledgement}
This research has been conducted as part of the DELPHI project, which is funded by the European Union, under grant agreement No 101104263. However, views and opinions expressed are those of the author(s) only and do not necessarily reflect those of the European Union or the European Climate, Infrastructure and Environment Executive Agency (CINEA). Neither the European Union nor the granting authority can be held responsible for them.

%\bibliographystyle{ieeetr}
%\bibliography{mybibliography}
%


%\end{thebibliography}
\end{document}